\documentclass[onefignum,onetabnum]{siamonline190516}

\usepackage{lipsum}
\usepackage{amsfonts}
\usepackage{graphicx}
\usepackage{epstopdf}
\ifpdf
  \DeclareGraphicsExtensions{.eps,.pdf,.png,.jpg}
\else
  \DeclareGraphicsExtensions{.eps}
\fi

\usepackage{enumitem}
\setlist[enumerate]{leftmargin=.5in}
\setlist[itemize]{leftmargin=.5in}

\newsiamremark{remark}{Remark}
\newsiamremark{hypothesis}{Hypothesis}
\crefname{hypothesis}{Hypothesis}{Hypotheses}
\newsiamthm{claim}{Claim}
\newsiamremark{example}{Example}

\headers{Train Like a (Var)Pro}{E.~Newman, L.~Ruthotto, J.~Hart, and B.~van Bloemen Waanders}

\title{Train Like a (Var)Pro: Efficient Training of Neural Networks with Variable Projection\thanks{Submitted to the editors on \today.
\funding{LR's and EN's work was supported by Sandia Contract 2003941 and NSF DMS 1751636.}}}

\author{Elizabeth Newman\thanks{Department of Mathematics, Emory University, Atlanta, GA, USA  
  (\email{elizabeth.newman@emory.edu}, \url{http://math.emory.edu/\~enewma5/}).}
\and Lars Ruthotto\thanks{Departments of Mathematics and Computer Science, Emory University, Atlanta, GA, USA,  
  (\email{lruthotto@emory.edu}, \url{http://math.emory.edu/\~lruthot/})
}
\and Joseph Hart\thanks{Sandia National Laboratories, Albuquerque, NM, USA}
\and Bart van Bloemen Waanders\thanks{Sandia National Laboratories, Albuquerque, NM, USA}}
\usepackage{amsopn}
\DeclareMathOperator{\diag}{diag}

\newcommand{\hf}{{\frac 12}}

\newcommand{\bfB}{{\bf B}}
\newcommand{\bfC}{{\bf C}}

\newcommand{\bfH}{{\bf H}}
\newcommand{\bfI}{{\bf I}}
\newcommand{\bfJ}{{\bf J}}
\newcommand{\bfK}{{\bf K}}

\newcommand{\bfQ}{{\bf Q}}
\newcommand{\bfR}{{\bf R}}

\newcommand{\bfU}{{\bf U}}
\newcommand{\bfV}{{\bf V}}
\newcommand{\bfW}{{\bf W}}

\newcommand{\bfY}{{\bf Y}}
\newcommand{\bfZ}{{\bf Z}}

\newcommand{\bfb}{{\bf b}}
\newcommand{\bfc}{{\bf c}}
\newcommand{\bfe}{{\bf e}}

\newcommand{\bfs}{{\bf s}}
\newcommand{\bfx}{{\bf x}}
\newcommand{\bfy}{ {\bf y}}
\newcommand{\bfu}{{\bf u}}

\newcommand{\bfv}{{\bf v}}
\newcommand{\bfw}{{\bf w}}

\newcommand{\bfg}{{\bf g}}
\newcommand{\bfz}{{\bf z}}

\newcommand{\bftheta}{{\boldsymbol \theta}}

\newcommand{\st}{\mathrm{s.t.}}

\newcommand{\R}{\ensuremath{\mathds{R}}}

\newcommand{\nFeatIn}{N_{\rm in}}
\newcommand{\nFeatOut}{N_{\rm out}}
\newcommand{\nTargets}{N_{\rm target}}
\newcommand{\nTarget}{N_{\rm target}}
\newcommand{\nWeights}{N_{\theta}}  % number of network weights

\newcommand{\weights}{\bftheta}  % network weights

\usepackage{comment}
\usepackage{amssymb, amsfonts, amsmath, latexsym, dsfont}
\usepackage{tikz}
\usepackage{algorithm, algpseudocode, float}
\usepackage{color, colortbl}
\definecolor{rowGray}{gray}{0.85}
\usepackage[first=0,last=9]{lcg} % row color 
\usepackage{multirow, array}
\newdimen\iwidth
\newdimen\iheight

\DeclareMathOperator*{\argmin}{\text{arg min}}

\ifpdf
\hypersetup{
  pdftitle={Train Like a (Var)Pro: Efficient Training of Neural Networks with Variable Projection},
  pdfauthor={E. Newman, L. Ruthotto, J. Hart, B. van Bloemen Waanders}
}
\fi

\newcommand{\change}[1]{{\color{black} #1}}

\begin{document}

\maketitle
\begin{abstract}% cannot exceed 250 words
Deep neural networks (DNNs) have achieved state-of-the-art performance across a variety of traditional machine learning tasks, e.g., speech recognition, image classification, and segmentation.
The ability of DNNs to efficiently approximate high-dimensional functions has also motivated their use in scientific applications, e.g.,  to solve partial differential equations (PDE) and to generate surrogate models. 
In this paper, we consider the supervised training of DNNs, which arises in many of the above applications.
We focus on the central problem of optimizing the weights of the given DNN such that it accurately approximates the relation between observed input and target data.
%bvbw Devising effective solvers for this optimization problem is notoriously challenging, e.g., due to the large number of weights, non-convexity, data-scarcity, and non-trivial choice of hyperparameters.
Devising effective solvers for this optimization problem is notoriously challenging  due to the large number of weights, non-convexity, data-sparsity, and non-trivial choice of hyperparameters.
To solve the optimization problem more efficiently, we propose the use of variable projection (VarPro), a method originally designed for separable nonlinear least-squares problems.
Our main contribution is the Gauss-Newton VarPro method (GNvpro) that extends the reach of the VarPro idea to non-quadratic objective functions, most notably, cross-entropy loss functions arising in classification.
These extensions make GNvpro applicable to all training problems that involve a DNN whose last layer is an affine mapping, which is common in many state-of-the-art architectures.
\change{In our four numerical experiments from surrogate modeling, segmentation, and classification GNvpro solves the optimization problem more efficiently than commonly-used stochastic gradient descent (SGD) schemes. Also, GNvpro finds solutions that generalize well, and in all but one example better than well-tuned SGD methods, to unseen data points.}
\end{abstract}


% ============================================================= %
\section{Introduction} % (fold)
\label{sec:introduction}
	
Deep neural networks (DNNs) have emerged as one of the most promising machine learning methods by achieving state-of-the-art performance in a wide range of tasks such as speech recognition~\cite{Hinton:2012acousticModeling}, image classification~\cite{Krizhevsky:2012alexnet}, and segmentation~\cite{ronneberger2015unet}; for a general introduction to DNNs, see~\cite{Goodfellow:2016wc}.  
In the last few years, these successes have also motivated the use of DNNs in scientific applications, for example, to solve partial differential equations (PDEs)~\cite{han2018solving,Raissi:2019hv,Sirignano:ik,weinan2018deep} and to generate PDE surrogate models~\cite{Tripathy:2018kk}.  

A key contributor to DNNs' success across these applications is their universal approximation properties~\cite{Cybenko1989} as well as recent advances in computational hardware, maturity of machine learning software,  and the availability of data.
However, in practice, training DNNs remains a challenging art that requires careful hyper-parameter tuning and architecture selection.  
Mathematical insights are critical to overcome some of these challenges and to create principled learning approaches. 
In this paper, we leverage methods and algorithms from computational science to design a reliable, practical tool to train DNNs efficiently and effectively.

DNNs are parameterized mappings between input-target pairs, $(\bfy, \bfc)$.
Learning means finding the weights of the mapping to achieve accurate approximations of the input-target relationships.
In supervised learning, the weights of the mapping are computed by minimizing an expected loss or discrepancy of $G(\bfy) \approx \bfc$ approximated using labeled training data.  
This optimization problem is difficult because the number of weights and examples is typically large, the objective function is non-convex, and the final weights must generalize beyond the final training data. 
The latter motivates the use of explicit (e.g., weight-decay or Tikhonov) or implicit (e.g., early stopping of iterative methods) regularization. 

Devising effective solvers for the optimization problem has received a lot of attention and two predominant iterative approaches have emerged: stochastic approximation schemes~\cite{RobbinsMonro1951}, such as stochastic gradient descent (SGD) (or variants like ADAM~\cite{kingma2014adam}), and stochastic average approximation (SAA) schemes~\cite{KleywegtEtAl2006}.
Each step of SGD approximates the gradient of the expected loss with a small, randomly-chosen batch of examples that are used to update the DNN weights.
Under suitable choices of the step size, also called the learning rate, each step reduces the expected loss and the iterations have been shown empirically to often converge to global minimizers that generalize well. 
In practice, however, it can be difficult to determine the most effective SGD variants and tune hyperparameters such as the learning rate. Furthermore, even if the scheme converges, the rate is often slow and the sequential nature of the method complicates parallel implementation.
Finally, it is non-trivial to incorporate curvature information~\cite{bottou2016optimization, Goldstein:2018lossLandscape, yao2020adahessian}.
An alternative to SGD approaches is minimizing a sample average approximation (SAA) of the objective function, which attempts to overcome some of the aforementioned disadvantages.
The accuracy of the average approximation improves with larger batch sizes and can be further increased by repeated sampling~\cite{Nemirovski:2009saa, KleywegtEtAl2006}.  
While the computational complexity of a step is proportional to the batch size, larger batches provide more potential for data parallelism. 
The resulting optimization problem can be solved using deterministic methods including inexact Newton schemes~\cite{Bollapragada_2018, OLearyRoseberry:2019vf, Xu_2020}. 
Particularly the examples in~\cite{OLearyRoseberry:2019vf} show the competitiveness of Newton-Krylov methods to common SGD approaches in terms of computational cost and generalization. 

Our goal is to further improve the efficiency of SAA methods by exploiting the structure of the DNN architecture.
 Many state-of-the-art DNNs~\cite{he2016deep, Raissi:2019hv, Lecun:1990mnist, Krizhevsky:2012alexnet, ronneberger2015unet} can be expressed in a separable form
 	\begin{align}\label{eq:splitFeaturesClassification}
 	G(\bfy, \bfW,\weights) = \bfW F(\bfy, \weights).
 	\end{align}
Here, $\bfW$ are the parameters of a linear transformation and $\weights$ parameterize the nonlinear feature extractor $F$. 
The key idea in this paper is to exploit the separable structure by eliminating the linear parameters through partial optimization.  \change{During training,} this reduces $G$ to
	\begin{align}\label{eq:splitFeaturesClassificationVarPro}
	G_{\rm red}(\bfy, \weights) = \bfW(\weights) F(\bfy, \weights),
	\end{align} 
where $\bfW(\weights)$ denotes the optimal linear transformation for the current feature extractor $F(\cdot, \weights)$.
For example, for least-squares loss functions, $\bfW(\weights)F(\bfy, \weights)$ is the projection of the target feature $\bfc$ onto the subspace spanned by the nonlinear feature extractor.
Hence, the resulting scheme is known as variable projection (VarPro); for the original works, see~\cite{GolubPereyra1973, Kaufman1975} and for excellent surveys refer to~\cite{GolubPereyra2003, OLearyRust2013}.

VarPro has been widely used to solve separable, nonlinear least-squares problems (see~\cite{GolubPereyra2003} and references therein), and its first applications to DNNs, to our best knowledge, can be attributed to~\cite{Sjoberg1997, Pereyra:2006jr}. 
These works train relatively shallow neural networks as function approximators and therefore use a regression loss function.
In addition to the obvious advantage of reducing the number of weights, these works also show that VarPro can create more accurate models and can accelerate the training process. 
The analysis in~\cite{Sjoberg1997} reveals that the faster rate of convergence can be explained by the improved conditioning of the optimization problem. 
Another advantage of VarPro is its ability to capture the coupling between the variables $\bfW$ and $\bftheta$, which is to be expected in our application.
In the presence of tight coupling between these blocks, VarPro can outperform block coordinate descent approaches (see, e.g.,~\cite{ChungEtAl2006} for a detailed comparison for an imaging problem). Using block coordinate descent to train neural networks has recently been proposed in~\cite{patel2020block,cyr2020robust}.

Our main contribution is the development of a Gauss-Newton-Krylov implementation of VarPro (GNvpro) that extends its use beyond least-squares functions, specifically to cross-entropy loss functions arising in classification.  
For cross-entropy loss functions, there is no closed-form expression for $\bfW(\weights)$, and therefore, to compute \cref{eq:splitFeaturesClassificationVarPro} efficiently and robustly, we \change{use a} trust region Newton scheme.
Due to our efficient implementation of this scheme the overhead of computing $\bfW(\weights)$ is negligible in practice and the computational cost of DNN training is dominated by forward and backward passes through the feature extractor.  
We compute the Jacobian of $\bfW(\weights)$, which is required in GNvpro, using implicit differentiation and accelerate the computation by re-using (low-rank) factorizations.  
As these two steps are independent of the choice of $F$, GNvpro is applicable to a general class of DNN architecture. 
To demonstrate its versatility, we experiment with the network architecture and as learning tasks consider \change{PDE surrogate modeling, segmentation, and classification}.  
Across those examples and as expected, GNvpro accelerates the convergence of the optimization problem compared to Gauss-Newton-Krylov schemes that train \cref{eq:splitFeaturesClassification} and SGD variants.   
Equally important, DNNs are trained that generalize well.  
GNvpro's ability to train a relatively small DNN to high accuracy is particularly attractive for surrogate modeling as the goal here is to reduce the computational cost of expensive PDE solves. 

\begin{figure}
\centering
\includegraphics[width=\textwidth]{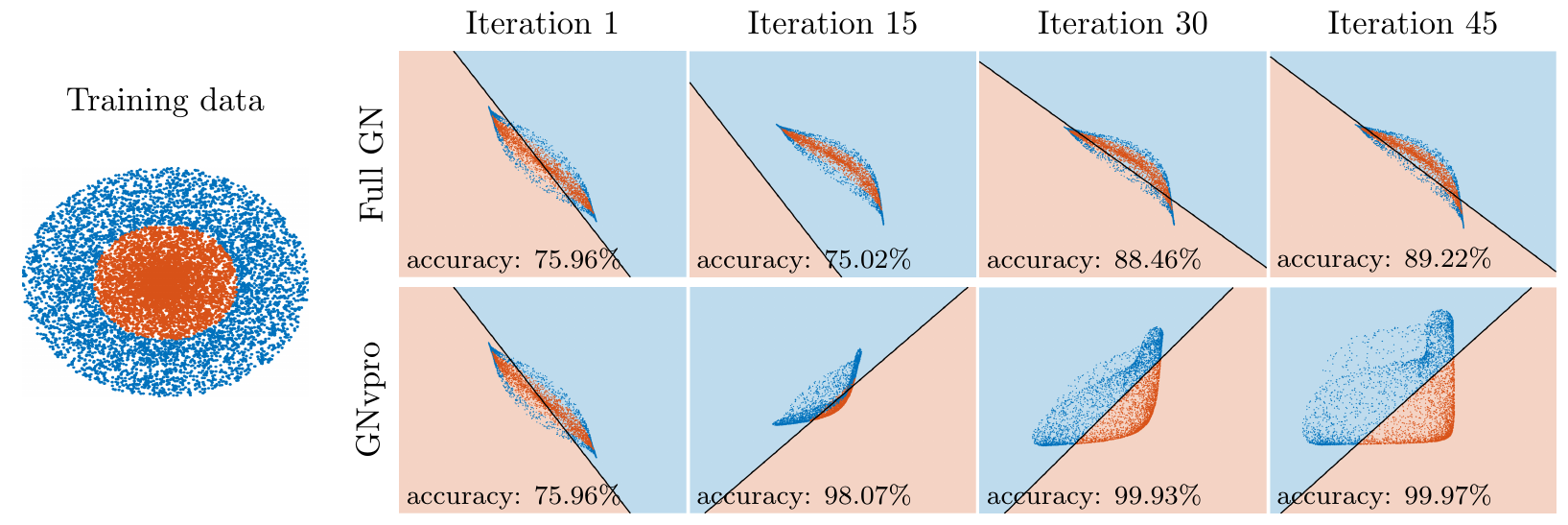}

\caption{ 
We compare the training performance of the full model (top row) and the VarPro-reduced model (bottom) given by \cref{eq:splitFeaturesClassification} and \cref{eq:splitFeaturesClassificationVarPro}, respectively, for a two-dimensional binary classification problem. %; for details, see \Cref{sub:circle}.
The training data consists of inputs $\bfy \in \mathbb R^2$ and are visualized in the left figure using red and blue dots based on their label $c \in \{0,1\}$.
In the other panels, we display the transformed training data $F(\bfy,\weights)$ at intermediate optimization iterations, the linear classifier $\bfW$ via a black line, and the resulting classification via the background shading. 
By design, each iteration of VarPro uses the optimal linear classifier for the transformed data and thus converges quicker and to a more accurate solution.}
\label{fig:hyperplanesCircle}

\end{figure}

Before deriving our algorithm, we establish a geometric intuition of VarPro and highlight the effectiveness of GNvpro through a simple binary classification example in \Cref{fig:hyperplanesCircle}. 
Our data consists of input-target pairs $(\bfy, c)$ where $\bfy \in \mathbb R^2$ are points lying in the Cartesian plane belonging to one of two classes:  contained in an ellipse ($c=0$) or outside the ellipse ($c=1$). 
We train a network with a three-layer feature extractor, written as the composition
	\begin{align*}
	F(\bfy, \weights) = \sigma(\bfK_2\sigma(\bfK_1\sigma(\bfK_0 \bfy + \bfb_0) + \bfb_1)+\bfb_2),
	\end{align*}
where the matrices are of size $\bfK_0\in \R^{4\times 2}$, $\bfK_1\in \R^{4\times 4}$, and $\bfK_2\in \R^{2\times 4}$, the vectors are of size $\bfb_0\in \R^4$, $\bfb_1\in \R^4$, and $\bfb_2\in \R^2$, and the activation function $\sigma(x) = \tanh(x)$ acts entry-wise.
For notational convenience, we let $\weights = (\bfK_0,\bfb_0, \bfK_1, \bfb_1, \bfK_2, \bfb_2)$ be a vector containing all of the trainable weights. 
We use our trust region Gauss-Newton-Krylov and Tikhonov regularization with and without VarPro to compare optimization strategies. 
As illustrated in \Cref{fig:hyperplanesCircle}, GNvpro improves the accuracy and efficiency and illustrates a near-optimal geometric evolution throughout the iteration history.

The paper is organized as follows.  In \Cref{sec:methods}, we derive GNvpro and its components by generalizing the VarPro idea to cross-entropy loss functions and in particular discuss the effective solution of the partial minimization problem arising in classification problems.
In \Cref{sec:numericalExperiments}, \change{we provide four numerical examples consisting of two least-squares regression problems, an image segmentation problem, and an image classification problem.  }
The regression problems seek to train a DNN surrogate models of the parameter-to-observable map  for two PDEs, Convection Diffusion Reaction and Direct Current Resistivity. The segmentation problem seeks to train a DNN to classify pixels that correspond to different materials or crops in a hyperspectral image. \change{As am image classification benchmark, we consider the CIFAR-10 problem, which consists of training a convolutional neural network that classifies natural images.} 
In \Cref{sec:conclusions}, we summarize our findings and discuss future directions.

% ============================================================= %
\section{Train like a (Var)Pro} % (fold)
\label{sec:methods}
In this section, we present GNvpro, our Gauss-Newton-Krylov implementation of variable projection (VarPro) that enables training separable DNN architectures~\cref{eq:splitFeaturesClassification} to solve regression and classification problems.  
Our algorithmic process is predicated on formulating the stochastic optimization problem as a sample average approximation. 
To implement VarPro, we separate the nonlinear and linear DNN components into outer and inner problems, respectively. 
At each training iteration, we solve the inner problem to eliminate $\bfW$ as in~\cref{eq:splitFeaturesClassificationVarPro}.
The type of loss function dictates the choice of the solver for the inner problem: in the case of a least-squares function, an SVD solves the inner problem and in the case for the cross-entropy function, we employ a trust region approach. 
To efficiently solve the outer problem, we linearize the reduced forward model~\cref{eq:splitFeaturesClassificationVarPro} and derive GNvpro as a generalization of Gauss-Newton to non-quadratic loss functions.

\subsection{VarPro for DNN Training}
\label{sub:supervisedLearning}

Let $(\bfy, \bfc)$ be an input-target pair where $\bfy$ belongs to the feature space $\mathcal{Y} \subset\R^{\nFeatIn}$ and $\bfc$ belongs to the target space \change{$\mathcal{C}\subset\R^{\nTarget}$}.
We denote the data set, the set of all input-target pairs, as $\mathcal{D} \subset \mathcal{Y}\times \mathcal{C}$, and assume that it is large or even infinite.
In this work, we consider DNN models of the form \cref{eq:splitFeaturesClassification}, which consist of two components: a nonlinear, parametrized feature extractor $F: \mathcal{Y} \times \R^{\nWeights} \to \R^{\nFeatOut}$ and a final linear transformation $\bfW \in \R^{\nTargets \times \nFeatOut}$. 
We denote the weights of the feature extraction $F$ by $\weights\in \R^{\nWeights}$.

In a supervised setting, we use the labeled data in $\mathcal{D}$ to train the weights of the model. We pose this as the stochastic optimization problem
\begin{align}\label{eq:stochasticObj}
\min_{\bfW, \weights} \mathbb{E}[L(\bfW F(\bfy, \weights), \bfc)] + R(\weights) + S(\bfW),
\end{align}
where the loss function $L: \R^{\nTargets} \times\mathcal{C} \to \R$ is convex in its first argument, and
$R: \R^{\nWeights} \to \R$ and $S:\R^{\nTargets \times \nFeatOut} \to \R$ are convex regularizers.  
In this work, we choose the Tikhonov regularizers $R(\weights) = \frac{\alpha_1}{2} \|\bfB\weights\|^2$ (where $\bfB$ is a user-defined regularization operator) and $S(\bfW) = \frac{\alpha_2}{2} \|\bfW\|^2_{\rm F}$ (where $\|\cdot \|_F$ is the Frobenius norm)  
 with fixed parameters $\alpha_1,\alpha_2>0$, which are called weight-decay in the machine learning community.  
The expectation is over the uniform distribution defined by the input-output pairs in $\mathcal{D}$. 
We list some  loss functions that are commonly-used in DNN training and supported by our framework in  \Cref{tab:lossFunctions}.

% loss function table
\begin{table}
\centering
%\caption[Loss functions supported in our framework for training DNNs to solve function approximation (least-squares) and classification tasks (logistic, multinomial).
%The first argument of the loss functions is the output of the network, i.e., $\bfx = \bfW F(\bfy, \weights)$. 
%The softmax function $h(\bfx) = \exp(\bfx) / (1 + \bfe^\top \bfx)$ transforms the output into a vector of probabilities where the $i^{th}$ value is the probability of belonging to class $i$.  
%In the one-dimensional case, the softmax function is equivalent to the sigmoid function.]
%{Loss functions supported in our framework for training DNNs to solve function approximation (least-squares) and classification tasks (logistic, multinomial).
%The first argument of the loss functions is the output of the network, i.e., $\bfx = \bfW F(\bfy, \weights)$. 
%The softmax function $h(\bfx) = \exp(\bfx) / (1 + \bfe^\top \bfx)$ transforms the output into a vector of probabilities where the $i^{th}$ value is the probability of belonging to class $i$.  
%In the one-dimensional case, the softmax function is equivalent to the sigmoid function\footnotemark[1].}

\caption{Loss functions supported in our framework for training DNNs to solve function approximation (least-squares) and classification tasks (logistic, multinomial).
The first argument of the loss functions is the output of the network, i.e., $\bfx = \bfW F(\bfy, \weights)$. 
The softmax function $h(\bfx) = \exp(\bfx) / (1 + \bfe^\top \bfx)$ transforms the output into a vector of probabilities where the $i^{th}$ value is the probability of belonging to class $i$.  
In the one-dimensional case, the softmax function is equivalent to the sigmoid function.  
The unit simplex $\Delta^{N_{\rm target}}$ contains vectors with non-negative entries that sum to $1$ (i.e., vectors of probabilities).
}

\label{tab:lossFunctions}

\begin{tabular}{ll}
Least Squares & \multirow{2}{*}{$L_{\rm ls}(\bfx, \bfc)= \|\bfx - \bfc\|_2^2$}\\
$L_{\rm ls}: \R^{\nTargets} \times \R^{\nTargets} \to \R$ \\[0.5em]
\hline\\[-0.75em]
Logistic Regression& \multirow{2}{*}{$L_{\rm log}(x, c)=-c \log h(x) -(1-c) \log(1 - h(x))$}\\
$L_{\rm log}: \R \times \{0,1\} \to \R$ \\[0.5em]
\hline\\[-0.75em]
Multinomial Regression&  \multirow{2}{*}{\change{$L_{\rm mul}(\bfx, \bfc)=-\bfc^\top \log h(\bfx)$}}\\
$L_{\rm mul}: \R^{\nTargets} \times \Delta^{\nTargets} \to \R$ 
\end{tabular}
\end{table}

We turn \cref{eq:stochasticObj} into a deterministic problem by replacing the expected loss with its sample average approximation (SAA)~(see also \cite{KleywegtEtAl2006,Nemirovski:2009saa,Kim:2015saa})
	\begin{align}\label{eq:fullObj}
	\min_{\bfW, \weights} \Phi(\bfW, \weights) 
		\equiv \frac{1}{|\mathcal{T}|}\sum_{(\bfy, \bfc) \in \mathcal{T}} L(\bfW F(\bfy,\weights), \bfc) + R(\weights) + S(\bfW). 
	\end{align} % add subsampled
	Here, we partition the labeled data into a training set $\mathcal{D}_{\rm train}$, a validation set $\mathcal{D}_{\rm val}$, and a test set $\mathcal{D}_{\rm test}$, and choose the batch $\mathcal{T}$ from the training set, i.e., $\mathcal{T} \subset \mathcal{D}_{\rm train} \subset \mathcal{D}$. 
	The validation dataset is used to inform the network design and calibrate the regularization parameters and the test dataset is used to gauge how well the final model generalizes.

	In SAA methods, generalization can be achieved by choosing a sufficiently large batch size $|\mathcal{T}|$, as this closes the match between \eqref{eq:stochasticObj} and \eqref{eq:fullObj}.
	\change{When the required batch size becomes prohibitively large, one can solve a sequence of instances of the deterministic problem with different batches~\cite{KleywegtEtAl2006}.  }
	\change{This subsampling becomes necessary, e.g., when the entire training dataset cannot be stored in memory, which is typically the case in imaging problems due to the limited storage provided by GPUs. 
	Key differences between this stochastic variant of the SAA method and SGD methods is that batch sizes remain larger and typically significantly fewer optimization steps are needed. }

	Although the optimization problem~\eqref{eq:fullObj} can  be solved using a variety of deterministic numerical optimization methods, their effectiveness in practice can be disappointing (see top row of~\Cref{fig:hyperplanesCircle} as an example). 
	Challenges include the non-convexity of the objective function in $\bftheta$, immense computational costs in the large-scale setting (i.e., when the number of examples and the number of weights are large), and the coupling between the linear variable $\bfW$ and the DNN weights $\weights$ in~\cref{eq:splitFeaturesClassification}. 
	To derive a more effective approach, we exploit the separability of the DNN model and the convexification of the loss function by eliminating the variables of the linear part of the model to obtain the reduced optimization problem 
%We extend this idea to any objective function $\Phi$ that is convex in the first argument.   % so we can find a 
   	\begin{align} 
	\min_{\weights}  \Phi_{\rm red}(\weights)  &\equiv \Phi(\bfW(\weights), \weights) \label{eq:reducedObj}\\
	\text{ where } \;\bfW(\weights) &= \argmin_{\bfW}  \Phi(\bfW, \weights).\label{eq:Wargmin}
	\end{align}
	Since we eliminated the weights associated with the linear part of the model, we call  $\Phi_{\rm red}: \R^{\nWeights} \to \R$ the reduced objective function.
	For the least-squares loss function, this approach is known as variable projection (VarPro) (see~\cite{GolubPereyra1973, Kaufman1975,GolubPereyra2003, OLearyRust2013} and also our discussion in~\Cref{sec:leastSquares}), but as we show below the approach extends to other convex loss functions. 
	One of the key practical benefits of VarPro is that the minimization of~\eqref{eq:reducedObj} with respect to $\weights$ implicitly accounts for the coupling with the eliminated variable $\bfW$, which can dramatically accelerate the convergence (see also the discussion and experiments in~\cite{ChungEtAl2006}).

The reduced objective function~\cref{eq:reducedObj} can be minimized using gradient-based optimization, without differentiating the inner problem~\cref{eq:Wargmin}. 
In other words, the gradients of \cref{eq:reducedObj} and \cref{eq:fullObj} with respect to $\weights$ are equal, as can be verified via	
	\begin{align}\label{eq:varproGrad}
	\change{\nabla_{\weights}\Phi_{\rm red}(\weights) =
		J_{\weights}\bfw(\weights)^\top \nabla_{\bfw}\Phi(\bfW(\weights), \weights) + \nabla_{\weights}\Phi(\bfW(\weights), \weights)
		=\nabla_{\weights}\Phi(\bfW(\weights), \weights),}
	\end{align}
	\change{where $\bfw(\bftheta) = \text{vec}(\bfW(\bftheta))\in \R^{\nTarget \nFeatOut}$ (the operator $\text{vec}(\cdot)$ reshapes a matrix into a vector) and $J_{\weights}\bfw(\weights)$ is the Jacobian of $\bfw(\bftheta)$.}  
	This observation also implies that gradients can become inaccurate when the inner problem is not solved to sufficient accuracy;  \change{for numerical evidence, see~\cref{fig:subproblemAccuracy} in~\cref{app:innerProblem}.} 
	
	For the loss functions considered in this paper, the inner problem~\cref{eq:Wargmin} is smooth, strictly convex, and of moderate size. 
	Thus, it can be solved efficiently using convex optimization schemes. 
	The method by which we solve the inner optimization problem varies based on the loss function.  
For regression tasks, we solve the resulting least-squares problem directly using the singular value decomposition (see details in \Cref{sec:leastSquares}) and for the cross-entropy loss functions we solve the convex program iteratively using a Newton-Krylov trust region scheme (see details in \Cref{sec:crossEntropy}).

The computational cost of solving~\cref{eq:Wargmin} depends on the number of examples $|\mathcal{T}|$, the width of the network $\nFeatOut$, and the number of output features $\nTarget$. 
As $\nFeatOut$ and $\nTarget$ are typically on the order of tens or hundreds, even for large-scale learning tasks and due to our efficient implementations described in the following sections, the cost of solving the convex inner problems~\cref{eq:Wargmin} is minimal in practice.
% computational cost depends on choice of loss function
Importantly, these costs grow independent of the depth of the DNN used as a nonlinear feature extractor $F$.  
Hence, the cost of evaluating the reduced objective function in~\cref{eq:reducedObj} and its gradient~\cref{eq:varproGrad}  depends on the complexity of $F$.
As the depth of $F$ grows, the forward and backward passes through the network dominate the training costs  and the overhead of solving~\cref{eq:reducedObj} becomes negligible.  
  
We provide a side-by-side comparison of evaluating the full and reduced objective functions in \Cref{alg:wrapper}.  
The substantial difference is Line~\ref{alg_step:Wargmin} in which we solve for the optimal $\bfW(\weights)$ based on the current weights to evaluate the reduced objective. 
Observe that the architecture of the feature extractor is arbitrary, which allows us to experiment with a wide range of architectures, particularly deep networks (see~\cite{Sjoberg1997} for experiments with single-layer neural networks).
  
 \begin{algorithm}[t]
\caption{Evaluating objective functions}
\label{alg:wrapper}

\begin{algorithmic}[1]
\Statex \textbf{Inputs:} Batch $\mathcal{T}=\{(\bfy_i,\bfc_i) \ : \ i=1,2,\ldots, |\mathcal{T}|\}$,  feature extractor $F$,  weights $\weights$, and, $\bfW$  (only needed when not using VarPro)
%\Statex
\Statex \emph{Forward propagate}
\For{$i=1,\dots, |\mathcal{T}|$}
\State $\bfz_i(\weights) = F(\bfy_i, \weights)$
\EndFor
\Statex
%\For{$i=1,\dots, |\mathcal{T}|$}
\Statex \hspace{-\algorithmicindent}\hspace{-0.2cm}
	\begin{minipage}{0.5\textwidth} 
		\begin{algorithmic}[1]
		\makeatletter
		\setcounter{ALG@line}{3}
		\makeatother
		
		\Statex \emph{Evaluate $\Phi$ (no VarPro)}
		\State --------------------------
		\State $\Phi_{\rm c} = \Phi(\bfW, \weights)$
		\State $\nabla \Phi_{\rm c} = \nabla_{(\bfW, \weights)} \Phi(\bfW, \weights)$
		\State $J_{\rm c} = J_{(\bfW, \weights)}\left( \frac{1}{|\mathcal{T}|}\sum_{i=1}^{|\mathcal{T}|}\bfW \bfz_i(\weights)\right)$
		\end{algorithmic}
	\end{minipage}
	\vline
	\begin{minipage}{0.425\textwidth}
		\begin{algorithmic}[1]
		\makeatletter
		\setcounter{ALG@line}{3}
		\makeatother
		
		\Statex \emph{Evaluate $\Phi_{\rm red}$ (VarPro)}
		\State $\bfW(\weights) =  \argmin_{\bfW} \Phi(\bfW, \weights)$ \label{alg_step:Wargmin}%\Comment{VarPro step}
		\State $\Phi_{\rm c} =   \Phi_{\rm red}(\weights) \equiv \Phi(\bfW(\weights), \weights)$
		\State $\nabla \Phi_{\rm c} = \nabla_\weights \Phi_{\rm red}(\weights)$ \label{alg_step:grad}
		\State $J_{\rm c}= J_{\weights}\left( \frac{1}{|\mathcal{T}|}\sum_{i=1}^{|\mathcal{T}|}\bfW(\weights) \bfz_i(\weights)\right)$
		\end{algorithmic}
	\end{minipage}
\Statex
\State return $\Phi_{\rm c}$, $\nabla \Phi_{\rm c}$, and, for GNvpro, $J_{\rm c}$

\end{algorithmic}

\end{algorithm}

% ----------------------------------------------------------------------------------------------------------- %	
\subsection{Efficient Implementation for the Least-Squares Loss}
\label{sec:leastSquares}

For the least-squares loss function, the inner problem~\cref{eq:Wargmin} is a linear regression problem that we solve directly using the singular value decomposition (SVD).
For concreteness, we now outline our implementation for this important special case.
Our final scheme is equivalent to the ones derived in ~\cite{OLearyRust2013,GolubPereyra2003}.
For ease of notation, we arrange the input and target vectors of the current batch $\mathcal{T} \subset \mathcal{D}_{\rm train}$ column-wise in the matrices  $\bfY \in \R^{\nFeatIn \times |\mathcal{T}|}$ and $\bfC\in\R^{\nTargets \times |\mathcal{T}|}$, respectively.
Similarly, let the $i$-th column of $\bfZ(\weights) = F(\bfY, \weights) \in \R^{\nFeatOut\times |\mathcal{T}|}$ correspond to output feature $F(\bfy_i, \weights)$, where $\bfy_i$ is the $i-$th column of $\bfY$. 
Using this notation and assuming weight decay regularization, we can write the reduced regression problem compactly as
\begin{align}
\min_{\weights} \Phi_{\rm ls}(\weights) 
	&\equiv \tfrac{1}{2|\mathcal{T}|}\| \bfW(\weights) \bfZ(\weights) - \bfC \|_F^2 
	+ \tfrac{\alpha_1}{2} \|\weights\|_2^2 + \tfrac{\alpha_2}{2}\|\bfW(\weights)\|_F^2 \label{eq:reducedObjLS} \\
	\st \; \bfW(\weights) &= \argmin_{\bfW} \tfrac{1}{2|\mathcal{T}|}\| \bfW \bfZ(\weights) - \bfC \|_F^2 
	+ \tfrac{\alpha_2}{2}\|\bfW\|_F^2. \label{eq:WargminLS}
\end{align}
We note that the inner problem is separable by rows and hence the  $\nTarget$ rows of $\bfW(\weights)$ can be computed independently and in parallel.
We solve the inner optimization problem efficiently and stably using the (reduced) SVD~\cite{OLearyRust2013, Kaufman1975} of the network outputs
	\begin{align}\label{eq:svdZ}
	\tfrac{1}{\sqrt{|\mathcal{T}|}}\bfZ(\weights) = \bfU{\bf{\Sigma}}\bfV^\top,
	\end{align}
where the columns of $\bfU \in \R^{\nFeatOut\times\nFeatOut}$ and $\bfV\in \R^{|\mathcal{T}| \times \nFeatOut}$ are orthonormal and ${\bf\Sigma} \in \R^{\nFeatOut\times\nFeatOut}$ is  diagonal.   
Here, we assume the number of samples is greater than the number of output features which is a reasonable assumption when optimizing with SAA methods.   

% ----------------------------------------------------------------------------------------------------------- %
\subsection{Efficient Implementation for Cross-Entropy Loss Functions}
\label{sec:crossEntropy}

For the cross-entropy loss functions used in logistic and multinomial regression tasks, the inner problem~\cref{eq:Wargmin} generally does not admit a closed-form solution. 
For our choice of the regularizer $S$, the elimination requires solving a smooth and strictly convex problem.  
Typically, this problem has no more than a few thousand variables. 
To efficiently solve this problem to high accuracy, we use the Newton-Krylov trust region method described below.
We recall that due to~\cref{eq:varproGrad}, the convergence of the outer optimization scheme crucially depends on the accuracy of the Newton scheme.

% iteration
To iteratively solve~\cref{eq:Wargmin}, we initialize the weights with $\bfW^{(0)}\equiv \change{\bfW_{\rm prev}}$ and a user-specified trust region radius $\Delta^{(0)}$. 
\change{The initialization of $\bfW^{(0)}$ is the solution obtained during previous outer iteration.}
In the $j$-th iteration, we compute the update to the weight by approximately solving the quadratic program
\begin{equation}\label{eq:trW}
	\min_{\delta\bfw} \nabla_{\bfw} \Phi(\bfW^{(j)},\bftheta)^\top \delta\bfw + \hf \delta\bfw^\top \nabla_{\bfw}^2 \Phi(\bfW^{(j)},\bftheta) \delta\bfw \; \text{ subject to }\; \|\delta\bfw\| \leq \Delta^{(j)}.
\end{equation}
Although the number of variables, $n_{\bfw}=N_{\rm target} N_{\rm out}$, is modest in most DNN training problems, we improve the efficiency of the overall scheme by projecting this subproblem onto the  $r$-dimensional Krylov subspace $\mathcal{K}_r(\nabla_{\bfw}^2 \Phi(\bfW^{(j)},\bftheta), \nabla_{\bfw} \Phi(\bfW^{(j)},\bftheta))$.
Using the Arnoldi method, this gives a low-rank factorization such that
\begin{equation}
	\label{eq:lowRank}
	 \bfQ_{r+1} \bfH_r  = \nabla_{\bfw}^2 \Phi(\bfW^{(j)},\bftheta) \bfQ_r,
\end{equation}
where the columns of $\bfQ_{r+1} \in \R^{n_{\bfW} \times (r+1)}$ are orthonormal, $\bfQ_r \in \R^{n_{\bfW} \times r}$ contains the first $r$ columns of $\bfQ_{r+1}$, and 
$\bfH_r \in \R^{(r+1)\times r}$ is an upper Hessenberg matrix.
As the Hessian is symmetric,  $\bfH_r$ should be tridiagonal; however, this may not be the case in practice due to roundoff errors arising in the computation of matrix-vector products with the Hessian.
The rank $r$ is chosen adaptively by stopping the Arnoldi scheme when the estimate of 
\begin{equation*}
	\min_{\bfz \in \R^r} \frac{\| \bfH_{r} \bfz - \beta \bfe_1\|}{\beta}, \quad \text{ where } \quad \beta = \|\nabla_{\bfw} \Phi(\bfW^{(j)},\bftheta)\|, \quad \bfe_1 = (1,0,0,\ldots)^\top \in \R^{r+1}
	 % \nabla_{\bfw}^2 \Phi(\bfW^{(j)},\bftheta) \ \bfQ_r \bfz + \nabla_{\bfw} \Phi(\bfW^{(j)},\bftheta)\|}{\|\nabla_{\bfw}^2 \Phi(\bfW^{(j)},\bftheta)\|}
\end{equation*}
falls below a user-defined tolerance or a user-specified maximum rank, $r_{\rm max}$, is reached. 
We note that, in exact arithmetic, this stopping criteria renders our scheme equivalent to GMRES~\cite[Sec. 6.5]{Saad2003} without restarting when the trust region constraint is inactive.

Using the low-rank factorization~\cref{eq:lowRank} and replacing the inequality constraints with a quadratic penalty, we obtain the approximate step $\delta\bfw = \bfQ_r \bfz^*(\lambda)$, where for $\lambda\geq 0$,  $\bfz^*(\lambda)\in\R^r$  is given by
\begin{equation*}
	\bfz^*(\lambda) = \argmin_{\bfz\in\R^r} \hf \| \bfH_{r} \bfz - \beta \bfe_1 \|^2 + \frac{\lambda}{2} \| \bfz\|^2.
\end{equation*}
If $\|\bfz^*(0)\| > \Delta^{(j)}$ (i.e., the trust region constraint is violated), we use {\sc Matlab}'s \texttt{fminsearch} with initial bracket $\left[0, \frac{\beta}{\Delta^{(j)}}\right]$ to find the smallest value of $\lambda$ such that the trust region constraint holds.
Finally, we obtain the trial step $\bfW_{\rm trial} = \bfW^{(j)} + \delta \bfW$.

% updating of Delta
To accept or reject the trial step and update the trust region radius, we compare the predicted and actual reduction achieved by $\bfW_{\rm trial}$ and follow the guidelines in~\cite[Sec. 3.3]{Kelley:1999vd}. 
We note that the low-rank factorization is only updated when $\bfW_{\rm trial}$ is accepted. 
% stopping
The Newton scheme is stopped when a maximum number of iteration is reached or when the scheme has converged, defined as 
\begin{equation*}
	\frac{\|\nabla_{\bfw} \Phi(\bfW^{(j+1)},\bftheta)\|}{\|\nabla_{\bfw} \Phi(\bfW^{(0)},\bftheta)\|} \leq \epsilon_{\rm rel}, \quad \text{ or } \quad \|\nabla_{\bfw} \Phi(\bfW^{(j+1)},\bftheta)\| \leq \epsilon_{\rm abs},
\end{equation*}
where we set the relative and absolute tolerances to $\epsilon_{\rm rel}= 10^{-10}$ and $\epsilon_{\rm abs}= 10^{-10}$, respectively.

% ----------------------------------------------------------------------------------------------------------- %
\subsection{GNvpro}
\label{sub:WJacobian}

To solve the reduced optimization problem \cref{eq:reducedObj} accurately and efficiently we now present our proposed Gauss-Newton-Krylov implementation of VarPro, GNvpro.
We use the trust region scheme described in the previous section to iteratively update the weights $\bftheta$. 
Recall that the  outer problem generally is not convex. 

In the $k$-th iteration of GNvpro, we build the trust region subproblem using  the following approximation of the loss
\begin{equation}\label{eq:quadraticApproxLoss}
\begin{split}
L(G_{\rm red}(\bfy, \weights^{(k)}) + J_{\weights}G_{\rm red}\delta \weights, \bfc) 
	&\approx \\
	&\hspace{-2cm}L(G_{\rm red}(\bfy, \weights^{(k)}), \bfc) 
	+ \nabla L^\top J_{\weights}G_{\rm red} \delta \weights
	+ \frac{1}{2}\delta \weights^\top J_{\weights}G_{\rm red}^\top \nabla^2L J_{\weights}G_{\rm red} \delta \weights,
\end{split}
\end{equation}
where $\nabla L$ and $\nabla^2 L$ are, respectively, the gradient and the Hessian of $L(G_{\rm red}(\bfy, \weights^{(k)}), \bfc)$ with respect to the first argument, and  $J_{\weights}G_{\rm red}\in\R^{\nTarget\times\nWeights}$ is the Jacobian of the reduced model with respect to $\weights$
\begin{align}\label{eq:linearizeDNN}
G_{\rm red}(\bfy, \weights^{(k)} + \delta\weights) \approx G_{\rm red}(\bfy, \weights^{(k)}) + J_{\weights}G_{\rm red} \delta \weights.
\end{align}
We note that our approach is equivalent to the  classical Gauss-Newton method for nonlinear least-squares loss when we consider the regression loss.
GNvpro uses this approximation of the loss to formulate the trust region subproblem (compare to  ~\cref{eq:trW}) and uses  the same subspace projection, and trust region strategy as above albeit with  larger tolerances. 

We now derive the Jacobian of the VarPro reduced model, $J_{\weights}G_{\rm red}$, that appears in~\cref{eq:linearizeDNN}. 
	Applying the product rule to~\cref{eq:splitFeaturesClassificationVarPro}, the Jacobian of the reduced DNN model reads
	\begin{align}\label{eq:vecJac}
	J_{\weights}G_{\rm red}(\bfy, \weights) 
		= J_{\weights}(\bfW(\weights) F(\bfy, \weights)) 
		= \bfW(\weights) J_{\weights}F(\bfy, \weights) +  (F(\bfy, \weights)^\top \otimes \bfI_{\nTargets})J_{\weights}\bfw(\weights) 
	\end{align}
where $\otimes$ is the Kronecker product and $\bfI_{\nTargets}$ is the $\nTargets\times \nTargets$ identity matrix~\cite{MatrixCookbook}.
While the Jacobian of the network in the first term can be computed easily using automatic differentiation or using the chain rule, the second term is less trivial as it requires the Jacobian of the (vectorized) solution of the convex problem~\eqref{eq:Wargmin}.
We compute this term by implicitly differentiating the first order necessary condition $\nabla_{\bfw}\Phi(\bfW(\weights),\weights) = \bf 0$, which yields
	\begin{align}\label{eq:jacW}
	\bf0 &= \nabla_{\bfw}^2 \Phi(\bfW(\weights),\weights)J_{\weights}\bfw(\weights)
	+J_{\weights}\nabla_{\bfw}\Phi(\bfW(\weights),\weights) \\
	\Leftrightarrow 	J_{\weights}\bfw(\weights) &= -\nabla_{\bfw}^2 \Phi(\bfW(\weights),\weights)^{-1} J_{\weights}\nabla_{\bfw}\Phi(\bfW(\weights),\weights). \label{eq:implicitSolve}
	\end{align}
	Here, we note that the Hessian $\nabla_{\bfw}^2 \Phi(\bfW(\weights),\weights)$ is  positive definite due to the strict convexity of the objective function in $\bfW$.
	\change{See~\cref{app:jacobian} for a detailed derivation of $J_{\bftheta}\nabla_{\bfw} \Phi(\bfW(\bftheta),\bftheta)$.}

Finally, we note that matrix-vector products with $J_{\weights}G_{\rm red}(\bfy, \weights) $ and its transpose can be performed without the (costly) Kronecker product, that is, for vectors $\delta \weights\in\R^{\nWeights}$ and $\delta s \in \R^{\nFeatOut}$ we have
	\begin{align}
	 J_\weights G_{\rm red}\delta\weights
	 	&=  \bfW(\weights) (J_{\weights}F(\bfy, \weights)\delta\weights) + (J_{\weights}\bfw(\weights) \delta\weights) F(\bfy, \weights)
		\label{eq:J} \\
	 J_\weights  G_{\rm red}^\top \delta\bfs
	 	&=J_{\weights}F(\bfy, \weights)^\top(\bfW(\weights)^\top \delta\bfs)  + J_{\weights}\bfw(\weights)^\top (\text{vec}(\delta\bfs F(\bfy, \weights)^\top)).
		\label{eq:Jt}
	\end{align}
When~\cref{eq:implicitSolve} is solved efficiently, the cost of the matrix-vector products is dominated by the linearized forward or backward propagations through the feature extractor $F$.
We now provide efficient implementations of the Jacobian matrix-vector product for the least-squares and the cross-entropy loss, respectively.

\paragraph{Jacobian for Least-Squares Loss}
We can simplify the Jacobian $J_{\weights}\bfW(\weights)$ in \cref{eq:implicitSolve} using the SVD of $\bfZ(\weights)$ computed in~\cref{eq:svdZ}.
Denoting the residual by $\bfR(\weights) = \bfW(\weights)\bfZ(\weights) - \bfC$, we note that 
	\begin{align}
	\nabla_{\bfW}^2\Phi_{\rm ls}(\bfW(\weights),\weights)
		&= \bfU({\bf \Sigma}^2 + \alpha_2 \bfI_{\nFeatOut})^{-1} \bfU^\top\\
	J_{\weights}\nabla_{\bfW}\Phi_{\rm ls}(\bfW(\weights),\weights)
		&=
	\tfrac{1}{\sqrt{|\mathcal{T}|}}\bfW(\weights)J_{\weights}\bfZ(\weights)\bfV{\bf\Sigma}\bfU^\top + \tfrac{1}{|\mathcal{T}|}\bfR(\weights)J_{\weights}\bfZ(\weights)^\top
	\end{align} 
With these ingredients, the multiplication of a given vector $\delta \weights \in \R^{\nWeights}$ with the Jacobian, for example, simplifies to 
	\begin{align}\label{eq:jacWLS}
	J_{\weights} \bfW(\weights)\delta\weights 
		&= -\left(
			\tfrac{1}{\sqrt{|\mathcal{T}|}} \bfW(\weights) \delta\bfZ\bfV{\bf\Sigma}
			+\tfrac{1}{|\mathcal{T}|} \bfR(\weights)\delta\bfZ^\top \bfU
		\right)({\bf \Sigma}^2 + \alpha_2\bfI_{\nFeatOut})^{-1}\bfU^\top,
%	 \text{where } \delta\bfZ &=J_\weights \bfZ(\weights)\delta\weights \nolabel
	\end{align}
where we abbreviate the directional derivative of the feature extractor by $\delta\bfZ = J_\weights \bfZ(\weights)\delta\weights$ and cancel redundant terms.
Note that  multiplying the inverse Hessian from the right avoids forming the Kronecker product and enables parallel computation of all the rows.
The computation of matrix vector products with the transpose of the Jacobian is along the same lines.
%the variable $\bfW$ is on the left in \cref{eq:WargminLS} and we do not vectorize as we did in our original derivation.  
Efficient implementations of \cref{eq:jacWLS}  re-use the previously-computed SVD from \cref{eq:svdZ} to compute $J_{\weights}\bfW(\weights)$ and also evaluate the Jacobian of the feature extractor $J_\weights \bfZ(\weights)$ only once.

\paragraph{Jacobian for Cross-Entropy Loss}
Our implementation of GNvpro re-uses the low-rank decomposition of $\nabla_{\bfw}^2 \Phi(\bfW(\weights),\weights)$ computed in the last iteration of the inner trust region solve to accelerate matrix-vector products with the Jacobian \change{(see~\cref{app:jacobian} for more details)}. That is, we replace \cref{eq:implicitSolve}
by $- (\bfQ_r \bfH_r^{\dagger} \bfQ_{r+1}^\top)\ J_{\weights}\nabla_{\bfw}\Phi(\bfW(\weights),\weights)$, where $\bfH_r^{\dagger}$ is the Moore-Penrose inverse.
To increase the efficiency of computing matrix-vector products our implementation is based on performing a SVD of $\bfH_r$ and to derive a low-rank factorization of $\bfQ_r \bfH_r^{\dagger} \bfQ_{r+1}^\top$ and its transpose.
While it can be more accurate to use an iterative method in this step, we did not observe any negative effects on accuracy in our experiments.
Moreover, by this  choice we ensure that $ J_\weights  G_{\rm red}$ and $ J_\weights  G_{\rm red}^\top$ given by~\cref{eq:J} and \cref{eq:Jt}, respectively, are transposes of one another for any choice of $r$. 

% ============================================================= %
\section{Numerical Experiments}
\label{sec:numericalExperiments}

We provide numerical experiments that demonstrate the efficacy of training DNNs with GNvpro for  PDE surrogate modeling applications (\Cref{sub:surrogateModeling}), hyperspectral image segmentation (\Cref{sub:indianPines}), \change{and the CIFAR-10 image classification problem (\Cref{sec:cifar10}).}  
Across these tasks, which involve different loss functions and DNN architectures, solving the reduced problem~\cref{eq:reducedObj} using GNvpro leads to faster convergence and a more accurate model than solving the full problem~\cref{eq:fullObj} using a Gauss-Newton-Krylov method or solving the stochastic problem~\cref{eq:stochasticObj} using the SGD variant ADAM~\cite{kingma2014adam}.  % include BFGS? 

% ----------------------------------------------------------------------------------------------------------- %
% ----------------------------------------------------------------------------------------------------------- %
\subsection{Experimental Setup}
\label{sub:setup}
	
 In this section, we describe the DNN architectures used in our experiments, define our measure of computational complexity, and describe the optimization strategies we compare. 
We emphasize that GNvpro does not depend on the architectures   and refer to~\cite{Goodfellow:2016wc} for an overview of possible network designs.   
 
 \paragraph{Neural ODEs}
 In the following experiments, we consider feature extractors given by a neural Ordinary Differential Equation (Neural ODE); see~\cite{HaberRuthotto2017, E2017, ChenEtAl2018}.
 To this end, we define $F(\bfy,\bftheta)$ as a numerical approximation of $ \bfu(T)$ that satisfies the initial value problem
	\begin{align}\label{eq:resnet_layer}
	\bfu(t) = f(\bfu(t), \bfK(t), \bfb(t)) \quad \text{for}\quad t\in (0,T], \quad \bfu(0) = \sigma(\bfK_{\rm in} \bfy + \bfb_{\rm in})
	\end{align}
with an artificial time $t\in [0,T]$, activation function $\sigma(x)=\tanh(x)$, and  weights $\bfK : [0,T] \to \R^{\nFeatOut\times\nFeatOut}, \bfb : [0,T] \to \R^{\nFeatOut}, \bfK_{\rm in}\in  \R^{\nFeatOut\times\nFeatIn}$, and $\bfb_{\rm in} \in \R^{\nFeatOut}$.
For notational convenience we collectively denote the weights  as $\bftheta = (\bfK, \bfb, \bfK_{\rm in }, \bfb_{\rm in})$.

As shown in~\cite{HaberRuthotto2017}, the stability of the Neural ODE~\cref{eq:resnet_layer} depends both on the choice of weights, which are learned during training, and the  layer function $f$, which is specified prior to training. Here,  we use the anti-symmetric layer
\begin{equation*}
	f(\bfu, \bfK, \bfb) = \sigma\left((\bfK - \bfK^\top - \gamma \bfI) \bfu + \bfb\right),
\end{equation*}
with $\gamma=10^{-4}$. This design leads to a stable dynamic when $\bfK$ and $\bfb$ are constant in time. 

To train the Neural ODE, we first discretize its features $\bfu$ and weights, $\bfK$ and $\bfb$, on the nodes of an equidistant grid of $[0,T]$ with $d$ cells, and then optimize the discretized weights.
In our experiments, we use a fourth-order Runge-Kutta scheme.
The number of time steps, $d$, can also be seen as the depth of the network.
Hence, our approach is a discretize-optimize method, which has been shown to perform well also in ~\cite{GholamiEtAl2019, Onken2020DO}.

\paragraph{Optimization and Regularization}
We compare GNvpro to the SGD variant ADAM~\cite{kingma2014adam} and a standard Gauss-Newton method applied to the full optimization problem~\cref{eq:fullObj} and the quasi-Newton scheme L-BFGS (implemented following~\cite{NocedalWright2006}) applied to the VarPro problem~\cref{eq:reducedObj}; we call this scheme L-BFGSvpro for brevity.
We discuss the hyperparameters of each scheme (e.g., linear solvers, line search methods, learning rates) as well as regularization parameters in the respective sections as they are specific to each task. 

To exploit the fact that our learning problems are obtained by discretizing the Neural ODE~\cref{eq:resnet_layer}, we employ a multilevel training strategy similar to~\cite{HaberRuthotto2017,Chang:2017te,cyr2019multilevel}. 
The idea is to repeatedly solve the learning problem for a shallow-to-deep hierarchy of networks obtained by increasing the depth $d$ in the time discretization.
On the first level, we initialize $\bftheta$, the weights of the feature extractor $F$,  randomly. In our experiments with the full objective functions we initialized $\bfW$ by solving~\cref{eq:Wargmin} using the methods described in \Cref{sec:leastSquares} and  \Cref{sec:crossEntropy} for the initial network weights.
In our examples this has improved the performance over random initialization of $\bfW$ and also ensures that all schemes start with the same initial guess.
After approximately solving the problem on the first level, the final iterate of $\bftheta$ is prolongated in time using piecewise linear interpolation and used to initialize the optimization for the network architecture given by discretizing~\cref{eq:resnet_layer} with twice as many time steps.
This procedure is repeated until the desired depth of the network is reached. 

For the Neural ODE weights, the regularization operator $\bfB$ is a finite-difference, time discretization to promote smooth transitions between layers.  
For the other weights, we use an identity regularization operator.

\paragraph{Computational Costs}
We measure the computational cost of training neural networks with the various optimization schemes in terms of  work units, which we define as the sum of the number of forward and backward passes through the network \change{over all training samples}.  
\change{In other words, if we have a training data set of size $N$, a forward and backward pass for a single sample costs $2/N$ work units.}

These forward and backward passes are the most expensive computational steps in training, particularly for deep networks.  
A forward pass consists of forming the current function value $\Phi_{\rm c}$ in \Cref{alg:wrapper} or a forward application of the Jacobian $J_{\rm c}$. 
A backward pass consists of forming the current gradient $\nabla \Phi_{\rm c}$ in \Cref{alg:wrapper} or an 
application of $J_{\rm c}^\top$.  
\change{When using a multi-level training strategy, work units on the coarse level (fewer time steps/layers) are less expensive than the work units on the fine level (more time steps/layers). 
For simplicity, we ignore the dependency of the cost on the number of time steps, $d$, because in each experiment, the network architectures are fixed across optimization methods.
}
% For simplicity, we ignore the dependency of the cost on the number of time steps, $d$, which can vary in the multi-level training. 

The number of work units per step differs across the optimization methods. 
For example, each epoch of ADAM requires two work units: one to evaluate the objective function and one to compute its gradient for all examples in the batch (see left column in \Cref{alg:wrapper}).
Each iteration of L-BFGSvpro requires two work units to compute the objective function and its gradient plus a few additional work units to perform the line search.
The iterations of GNvpro require additional $2r$ work units to obtain the low-rank factorization of the approximate Hessian.
\change{The memory requirements of GNvpro is the same as full Gauss-Newton and the computational overhead of training with GNvpro is small.}

% ----------------------------------------------------------------------------------------------------------- %		
\subsection{Surrogate Modeling}
\label{sub:surrogateModeling}
		
In this section, we apply GNvpro to two datasets motivated by PDE surrogate modeling with DNNs. 
One of the key challenges for surrogate modeling consists of capturing the nonlinear features by efficiently manipulating the vast number of input parameters.  
Creating suitable surrogates that in turn can be used for a range of analysis tasks, such as uncertainty quantification and optimization, requires a necessary level of accuracy. However, when the underlying dynamics is nonlinear and the input to output mapping is nonconvex, the learning process becomes non-trivial.  

Our goal is to design a DNN that maps given parameters $\bfy$ to a set of observables $\bfc$, where $\bfy$ and $\bfc$ satisfy
\begin{align*}
	\bfc = \mathcal{P}u \quad \text{subject to} \quad \mathcal{A}(u;\bfy) = 0.
	\end{align*}
The function $u$ is the solution of the PDE and $\mathcal{A}$ is the PDE operator parameterized by $\bfy$.
%where $u:\Omega \to \R$ is the solution to the PDE in the domain $\Omega$, $\mathcal{A}$ is a PDE function that is potentially nonlinear in $u$, and $q:\Omega \to \R$ is the source. 
The linear operator $\mathcal{P}$ measures the solution $u$ at discrete points in the domain to obtain a set of observables $\bfc\in \R^{\nTargets}$.

%The goal of surrogate modeling is to find a small, efficient, accurate mapping from the input parameters $\bfy$ to the observables $\bfc$.  

%Our goal is to design a DNN that maps given parameters $\bfy$ to a set of observables $\bfc$, which are obtained by evaluating the solution of a PDE that models the underlying physics at a specified number of points.
To train the DNN surrogate, we minimize the least-squares regression loss.  
For these applications, it is reasonable to assume that the training data contains no noise or errors as it is generated using a fine mesh solver. 
Hence, an effective surrogate should tightly fit the training data and at the same time be efficient to evaluate.  
Note that we develop these surrogate models based solely on the parameter-observable pairs, independent of the underlying PDE.

In our experiments, we consider two examples: convection diffusion reaction (CDR), a commonly used PDE which models a variety of physical phenomena, and direct current resistivity (DCR), a inverse conductivity problem using a PDE model for electric potential. Both examples are summarized below and fully described in~\cref{appendix}. 
%The CDR equation is leveraged to model physical phenomena such as the transport and interaction species in subsurface and atmospheric flows, chemical interactions in combustion simulations, and the spread of wildfires 

The CDR equation has inspired mathematical models for studying problems in many fields, such as subsurface flows \cite{leAugeraud-JOTA-17,Li-AJCM-17}, atmospheric dispersion modeling \cite{Leelo-CEJG-14}, semi-conductor physics \cite{Junger-2013}, chemical vapor deposition \cite{Geiser-Polymer-13}, biomedical engineering \cite{Madzvamuse-BMB-02,Ferreira-PR-02}, population dynamics \cite{Yi-JDE-09,Baurmanna-JTB-07}, combustion \cite{Lucchesi-CTM-19}, and wildfire spread \cite{Grasso-2018}.  
%\joeynote{Bart, you and I have discussed some of these applications in the past. Did I summarize this correctly? Do you have any good references to add here?} 
In the CDR example we consider, the inputs $\bfy \in \mathbb R^{55}$ which parameterizes the reaction function and the targets $\bfc\in \R^{72}$ which are observations of the state $u$ at $6$ fixed spatial locations and $12$ instances of time. 
The data set consists of 800 pairs of parameters $(\bfy,\bfc)$ generated by repeatedly solving the PDE.
\change{We split the data into $400$ training samples ($50\%$ of available data), $200$ validation samples, and $200$ test samples. 
We are interested in being able to train small networks efficiently without a surplus of data -- a realistic setting in surrogate modeling where generating data is expensive.}

Direct Current Resistivity (DCR) seeks detect an object from indirect noisy measurements~\cite{SeidelLange2007,deymor}. The mathematical formulation is general and a similar instance of this problem also arises in modeling porous media flow with Darcy's Law~\cite{ChenHuanMaBook}. We illustrate surrogate modeling where the inputs $\bfy \in \R^{3}$ parameterize the depth, volume, and orientation of an ellipsoidal conductivity model and the targets $\bfc \in \R^{882}$ correspond to the observed electric potential differences. We generate $\change{10,000}$ samples of parameters and observables $(\bfy,\bfc)$.
\change{We split the data into $8,000$ training samples ($80\%$ of available data), $1,000$ validation samples, and $1,000$ test samples.}

%Direct Current Resistivity (DCR) is a type of inverse conductivity problem in which we detect an object from indirect noisy measurements; see, e.g., \cite{SeidelLange2007,deymor}. A similar instances of this problem arises in modeling porous media flow with Darcy's Law~\cite{ChenHuanMaBook}. We illustrate surrogate modeling where the inputs $\bfy \in \R^{3}$ parameterize the depth, volume, and orientation of an ellipsoidal region
%We generate the DCR data for a number of ellipsoidal conductivity models $m$ using the Finite Volume approach described in~\cite{Haber2015}.  
%The input features $\bfy_j\in \R^{3}$ are randomly chosen parameters to adjust ellipsoidal region in depth, volume, and orientation in the $xy$-plane of the conductivity region.  
%The output features $\bfc_j\in \R^{882}$ correspond to the observed electric potential differences in the $x$ and $y$ directions, measured at $21\times 21$ equally spaced locations on the subsurface.
%We consider $6000$ samples $(\bfy_j,\bfc_j)$, split into $4000$ training and $2000$ validation examples.  The same split is used for each optimization method.  
%Because we generated the data with a large dipole source, we subtract the mean from all the samples so we train on the distinguishing features rather than the dominant source.

%
For both examples, we train the Neural ODE model~\cref{eq:resnet_layer} with a width of $\nFeatOut=8$ and $\nFeatOut=16$ for the CDR and DCR examples, respectively, and a final time of $T=4$.  
Our multi-level approach consists of three steps whose respective number of time steps is $2$, $4$, and $8$.  
For GNvpro, we limit the rank of the Krylov subspace to $r_{\rm max}=20$  and use a relative residual tolerance of $10^{-2}$. 
\change{For the SAA methods, we use the entire training dataset at each iteration.}
For ADAM, we use the final depth of the network with a batch size of $2$ and \change{the recommended} learning rate of $10^{-3}$ from~\cite{kingma2014adam}.
For all optimization strategies, we use Tikhonov regularization with $\alpha_1 = \alpha_2 = 10^{-10}$ to prioritize fitting.

	\begin{table}
	\caption{Training, validation, and test results for convection diffusion reaction and DC Resistivity. 
	We report mean relative errors $\pm$ standard deviation where the smaller values indicate better fit. 
	\change{The displayed loss values are the final loss of each training method after a fixed number of work units (WU). 
	To strengthen the baseline for this comparison, we allocate twice as many work units to ADAM.}
	\label{tab:res}}
	
	\begin{center}
	\begin{tabular}{l|c|c|c|c}
		               & \change{WU}  &Training         & Validation & Test\\
	\hline
	\rowcolor{rowGray}
	\multicolumn{5}{c}{Convection Diffusion Reaction}       \\
	Full GN & \change{$600$}		& $0.0115	\pm 0.0055$ & $0.0138\pm 0.0075$ &  $0.0142\pm 0.0072$\\
	Full ADAM & \change{$1200$}      	& $0.0091\pm 0.0045$ & $0.0114\pm 0.0086$ &  $0.0116\pm 0.0072$\\ % no multilevel
	L-BFGSvpro &  \change{$600$} 	& $0.0171	\pm 0.0102$ &  $0.0219\pm 0.0127$ &  $0.0260\pm 0.0187$\\
	 GNvpro&   \change{$600$}  	&  $\bf 0.0045\pm 0.0021$ & $\bf 0.0057\pm0.0033$ &  $\bf 0.0060\pm 0.0032$\\ \hline
	\rowcolor{rowGray}
	 \multicolumn{5}{c}{DC Resistivity}       \\
	Full GN & \change{$1200$}     	& $0.3837	\pm 0.8384$ & $0.3650\pm0.6920$ &  $0.3517\pm 0.6638$\\
	% Full SGD      	& $0.2245	\pm 0.2819$ & $0.2153\pm 0.2426$ &  $0.2179\pm 0.2551$\\ % multilevel	
	Full ADAM & \change{$2400$}         	& $0.2994	\pm 0.3652$ & $0.2839\pm 0.2776$ &  $0.2904\pm 0.3256$\\ % no multilevel	
	L-BFGSvpro & \change{$1200$}      	&  $0.0786\pm 0.1878$ & $0.0763\pm 0.1887$ &  $0.0747\pm 0.1625$\\ 
	GNvpro & \change{$1200$}    & $\bf 0.0198	\pm 0.0311$ &  $\bf 0.0191\pm 0.0258$ &  $\bf 0.0199\pm 0.0322$\\\hline
	\end{tabular} 
	\end{center}
	\end{table}
	
\begin{figure}
	\includegraphics[width=\textwidth]{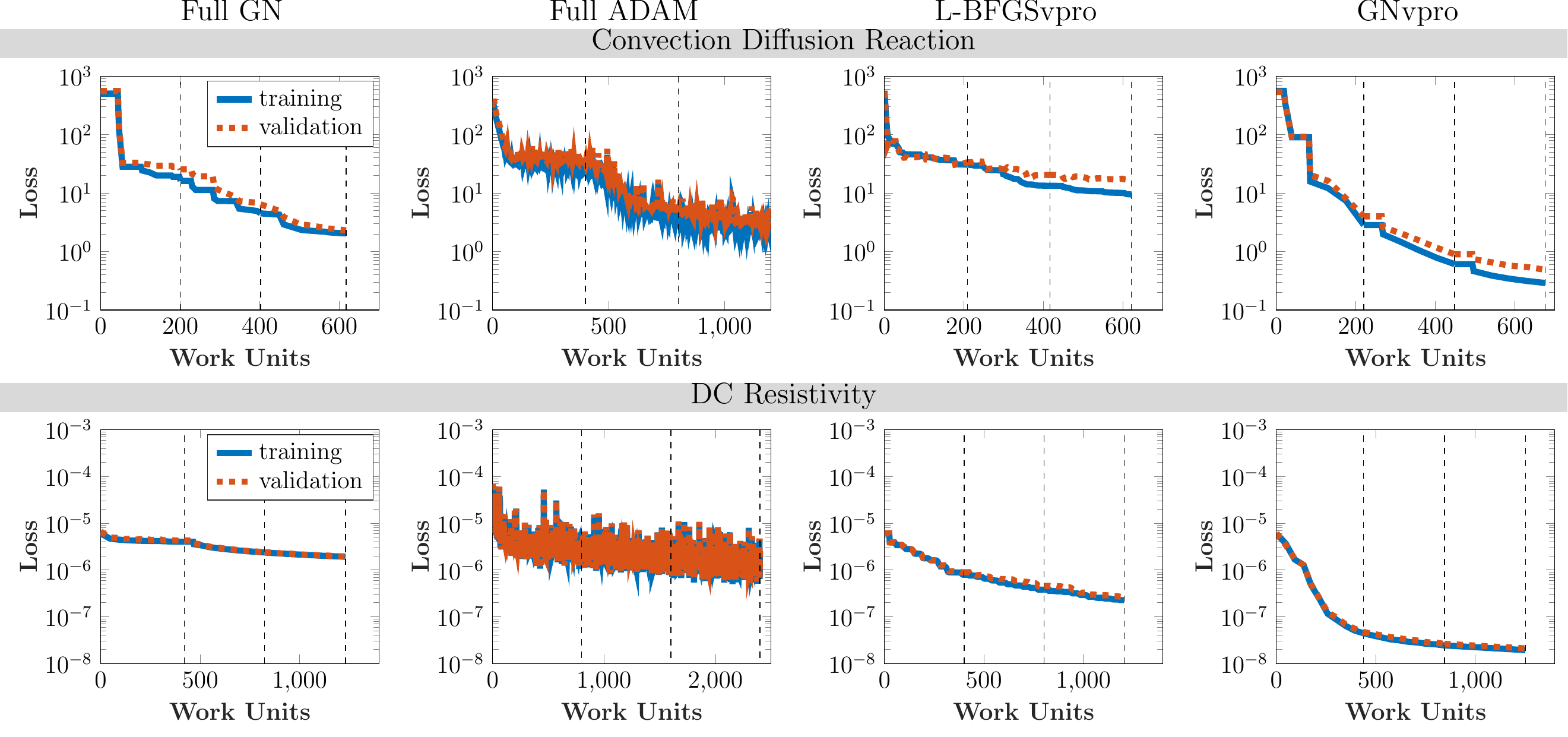}
	 \caption{%
	We plot the convergence of the loss in the CDR and DCR experiments for the various optimization strategies. 
	The weights of the Neural ODE are prolongated twice, indicated by the vertical dashed lines in each plot.  
	GNvpro outperforms all other approaches and achieves the best fit in the fewest number of work units whereas L-BFGSvpro achieves a moderate accuracy and the Gauss-Newton and ADAM schemes fail to achieve a sufficient accuracy. }
	 \label{fig:surrogateConvergence}
	\end{figure}

We report the CDR and DCR numerical results in \Cref{tab:res}. 
GNvpro produces the most accurate surrogate models for both the CDR and DCR experiments in \Cref{tab:res}.  
In the DCR experiment, GNvpro reduces the data fit by an additional order of magnitude compared to the Gauss-Newton and ADAM methods applied to the full objective function and also outperforms L-BFGSvpro.
As revealed by the convergence plots in \Cref{fig:surrogateConvergence}, GNvarpro achieves this level of accuracy in fewer work units.

\change{Generalization is particularly important in surrogate modeling and other scientific applications that require reliable solutions. 
Strikingly, the GNvpro solution also generalizes best, that is, it leads to the lowest validation and test errors in \Cref{tab:res} and most efficient reduction of the validation loss in \Cref{fig:surrogateConvergence}.}

\begin{figure}
	\centering
	\includegraphics[width=\textwidth]{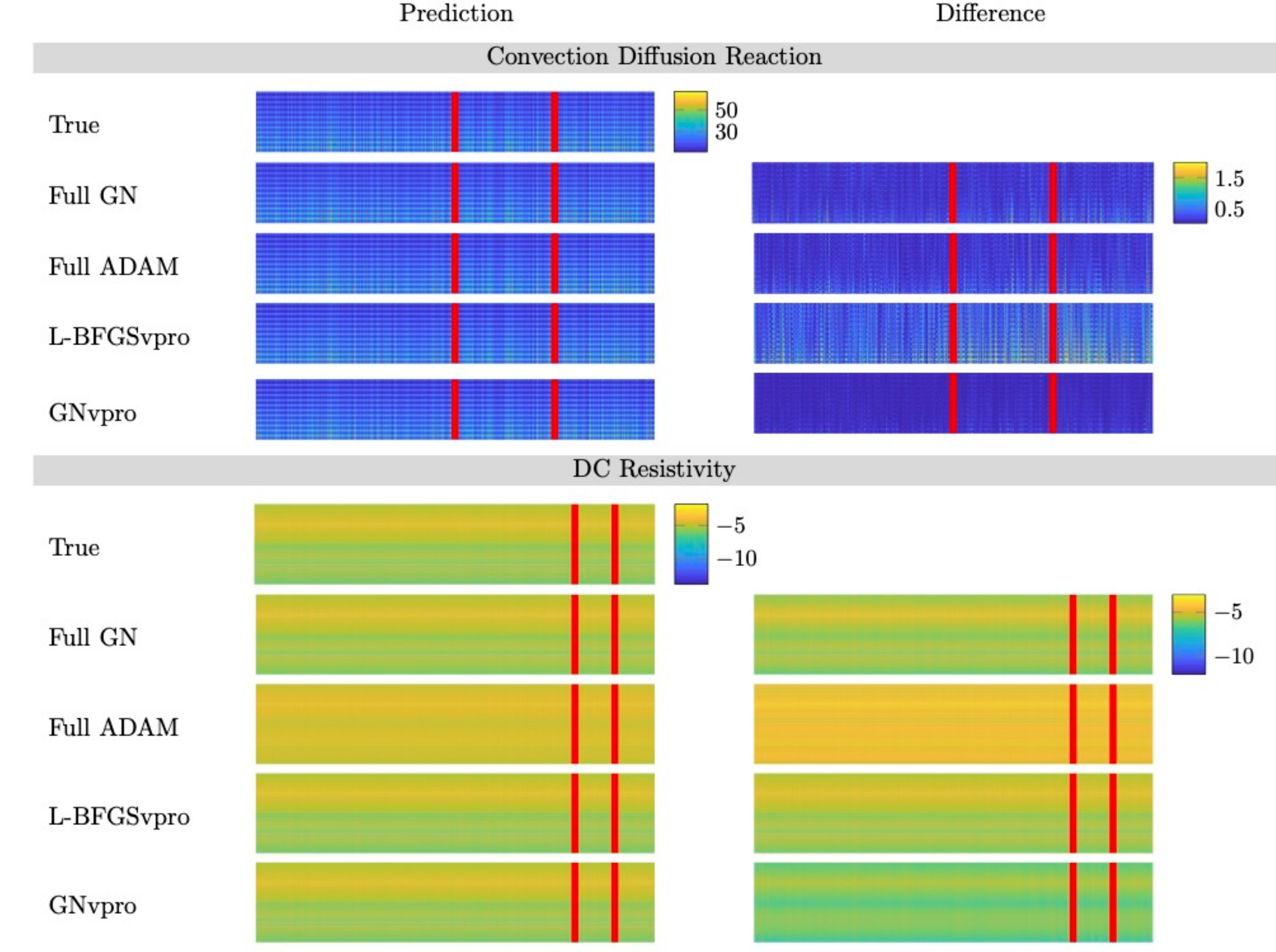}

	\caption{Global visualization of DNN surrogate model approximations for CDR and DCR data.  
	In the left images, each column of each image represents the output of the trained Neural ODE for different samples.  
	In the right images, we display the absolute difference between the predicted outputs and the true data.  
	To improve visibility of the DCR data, the absolute value of the data and their difference is used and intensities are scaled logarithmically. 
	The red lines divide the data into training, validation, and test sets, left-to-right.  
	From the difference images and numerical results in \Cref{tab:res}, the GNvpro is the most similar to the true, observed data.
	}
	\label{fig:surrogateModel}
	\end{figure}

To further support our numerical findings, we visualize the prediction obtained from the DNN surrogates for the CDR and DCR experiments in \Cref{fig:surrogateModel}.  
Each image in \Cref{fig:surrogateModel} contains all of the observables and the DNN approximations to the observables as columns of the matrix thereby providing a global visualization of the performance of our surrogate models.  
The width of each image is the total number of samples that we partition into training, validation, and test sets. 
The difference images are the most illuminating about the quality of approximation. 
We see that training with GNvpro produces difference images with values closest to zero, supporting our numerical findings in \Cref{tab:res}.

	% \surrogateGlobal

\begin{figure}
	\centering
	\includegraphics[width=\textwidth]{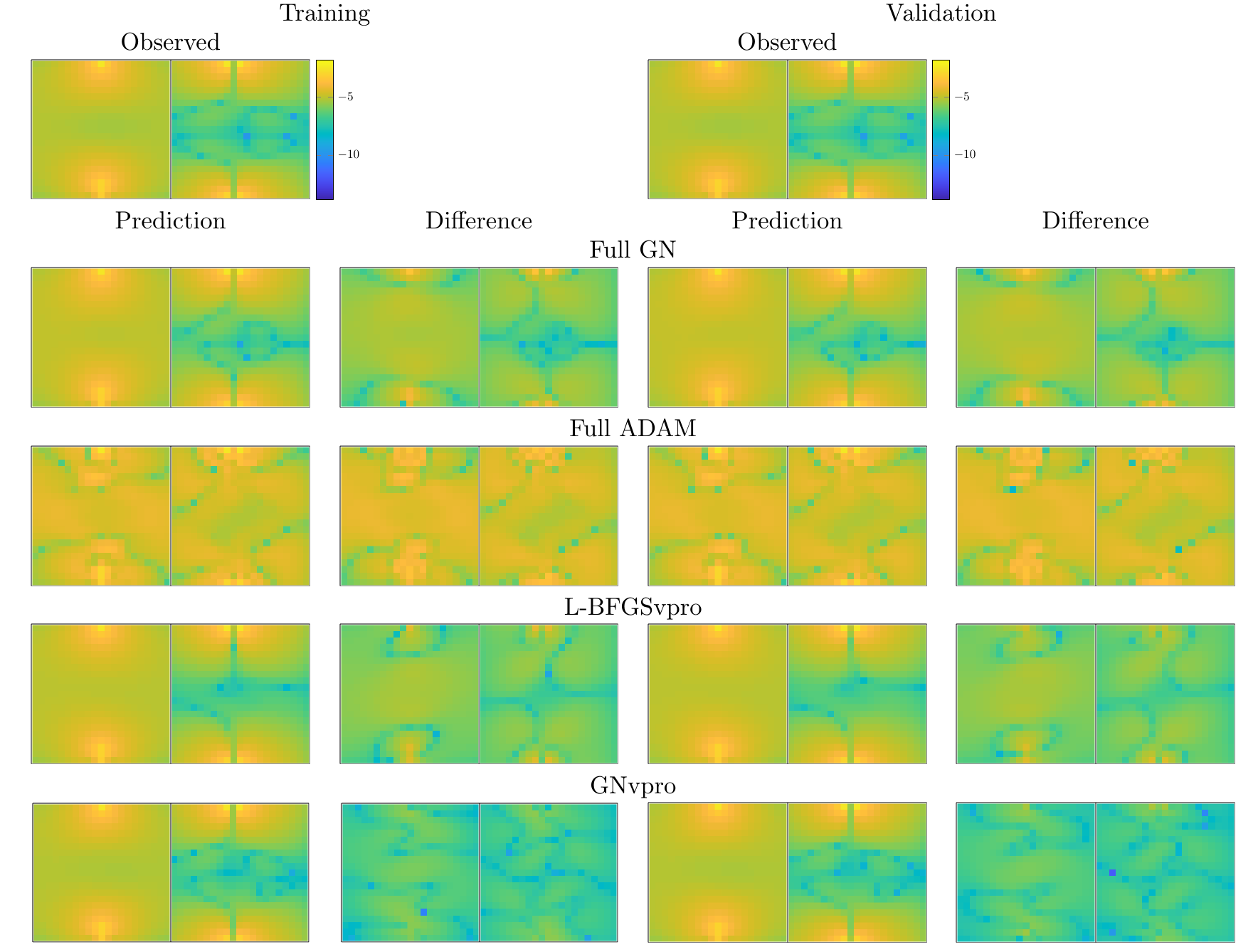}

		\caption{Local DCR surrogate model results.  
	The left two columns show an example from the training set as well as absolute errors of the prediction. 
	The right two columns show an example from the validation set. 
	% each example chosen to be the best VarPro GN approx
	To improve visibility, we show the absolute value of the data and scale all axes logarithmically. 
	%The same color axis is used in all corresponding plots to simplify comparison.  
	Each image is contains two $21\times 21$ blocks representing the difference in potential on the subsurface in the $x$-direction and $y$-direction, respectively.  
	%The large values at the top and bottom of each image represents the dipole source. 
	We notice that training the full models results in noisy approximations to the training and validation examples.  
	%However, training with VarPro retains some of the structure from 
	\label{fig:dcrLocal}}
	\end{figure}

For the DCR experiments, we can gain additional insight by visualizing the output $\bfc\in \R^{882}$ as two $21\times 21$ images for a few randomly chosen examples in \Cref{fig:dcrLocal}. 
As expected from the small loss values,  GNvpro  best captures the pattern of the observed, true data for both the training and validation examples. 
 L-BFGSvpro and Gauss-Newton retain some resemblance to the true data, including some of the ellipsoidal structure in the left blocks of each image. 
In comparison, the ADAM prediction does not capture the structure in the images, further demonstrating the competitiveness of GNvpro.

Together, these numerical results and visualizations demonstrate that training with GNvpro provides a more reliable PDE surrogate model that generalizes well and outperforms ADAM in terms of both convergence speed and accuracy.

\subsection{Indian Pines}
\label{sub:indianPines}
		
\begin{figure}
\centering
\includegraphics[width=0.9\textwidth]{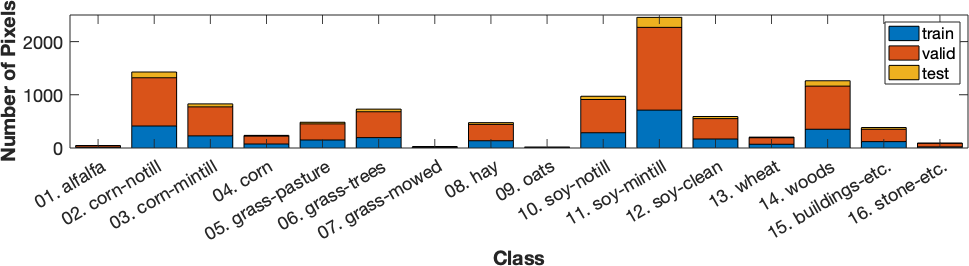}
\caption{Distribution of Indian Pines data.  
Each bar represents the total number of pixels belonging to each of the $16$ classes. 
The bars are split into training (blue) and validation (pixels).  
The smallest classes (1, 4, 7, 9, 13, 16) are split to have $10$ validation pixels, the remaining classes have $\lfloor 5000 / 16\rfloor = 312$ training pixels each.}
\label{fig:indianPinesDistribution}
\end{figure}

We demonstrate the ability of GNvpro to efficiently solve multinomial regression problems using the Indian Pine dataset~\cite{indianpinesdata}, a hyperspectral image dataset containing of electromagnetic images of agricultural crops.
The image data has $145\times 145$ pixels and $220$ spectral reflectance bands. 
The goal is to partition a hyperspectral image based on the crop or material in the given location.  
The $16$ different material types are provided in \Cref{fig:indianPinesDistribution}. 
We consider each pixel of data to be an input feature vector of length $220$.  
We randomly split the pixels into training, validation, and test datasets.
Achieving generalization of the DNN model is also complicated by the large variations of the number of pixels among the different crops. We visualize this class imbalance in \Cref{fig:indianPinesDistribution}.

\begin{figure}
	\centering
	
	\includegraphics[width=\textwidth] {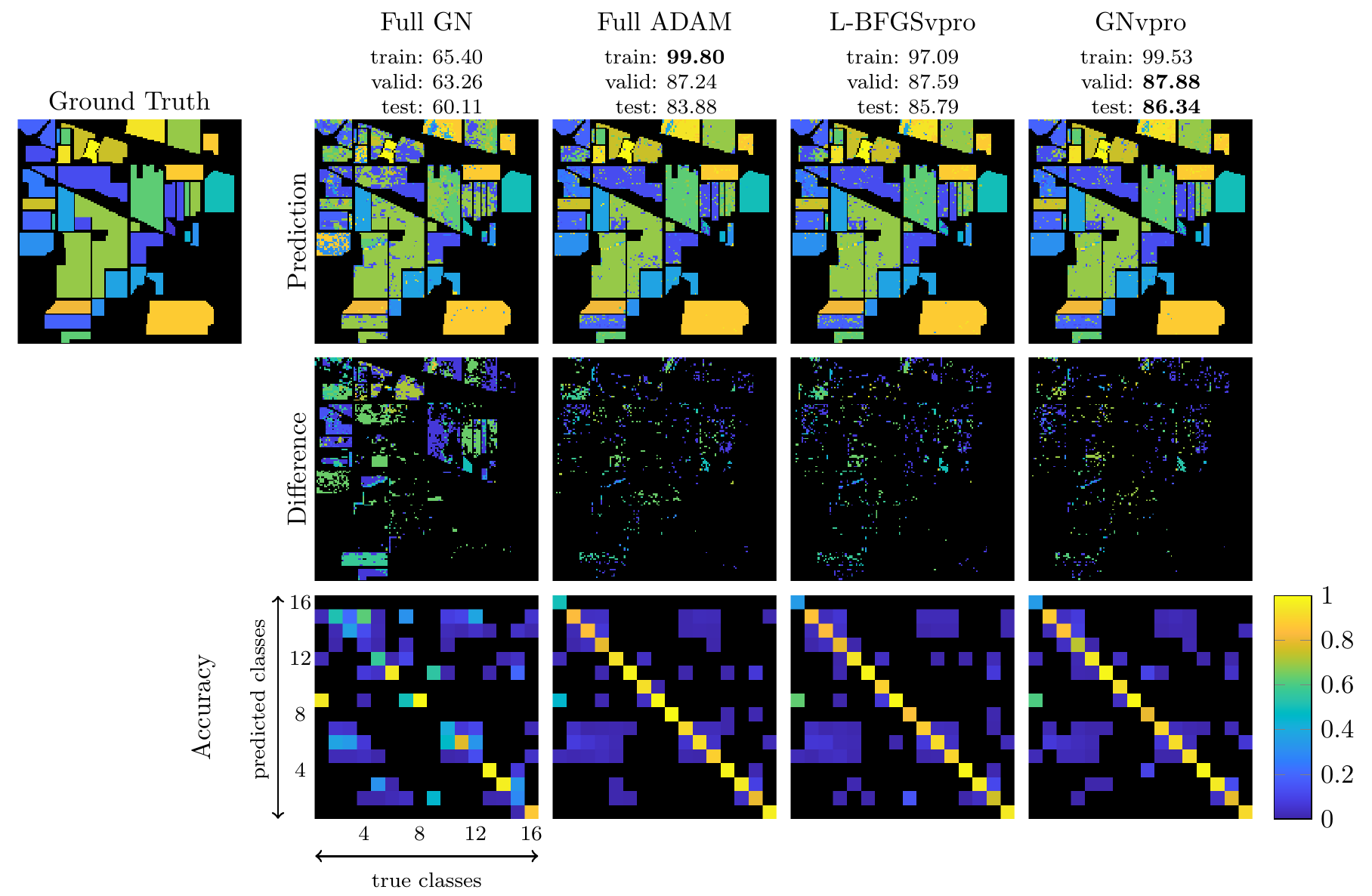}

	\caption{%
Comparison of optimization schemes for segmenting the Indian Pines hyperspectral dataset; each column corresponding to \change{Gauss-Newton, ADAM, L-BFGSvpro, and GNvpro}, respectively. 
The ``Prediction'' row contains the prediction maps where each color corresponds to a different class. 
We use the network that produced the highest validation accuracy to create the prediction. 
The ``Difference'' row contains the absolute difference between the prediction and the ground truth. 
The ``Accuracy'' row contains confusion matrices to visualize the accuracy per class.  
Each $(i,j)$-entry of the confusion matrices is the ratio of the number of times a pixel was classified as class $i$ that truly belonged to class $j$.  
Each column is normalized by the total number of pixels belonging to class $j$ from the true labeling, so the sum of every column is equal to $1$. 
Thus, each $(i,j)$-entry can be thought of as the probability that a predicted class $i$ belongs to class $j$.  
We want to see high probabilities along the diagonal entries.  
We plot the convergence of the accuracy and the loss for each optimization method for training and validation data.  
Note that we ran ADAM for twice as many work units as the other methods and kept the weights corresponding to the highest validation accuracy (black dot).
}
\label{fig:IndianPinesResults}
	\end{figure}

\begin{figure}
\centering
\includegraphics[width=\textwidth] {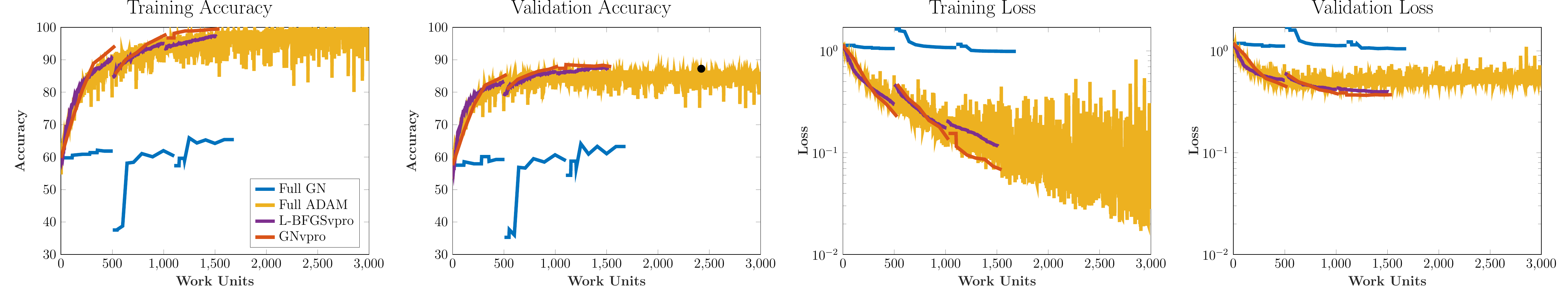}

	\caption{%
We plot the convergence of the accuracy and the loss for each optimization method for training and validation data.  
Note that we ran ADAM for twice as many work units as the other methods and kept the weights corresponding to the highest validation accuracy (black dot).
}
\label{fig:IndianPinesConvergence}
\end{figure}

We train the Neural ODE model~\cref{eq:resnet_layer} with a width of $\nFeatOut=32$ and a final time of $T=4$.  
Our multi-level approach consists of three steps whose respective number of time steps is $4, 8, $ and $16$.
For GNvpro, we limit the rank of the Krylov subspace to $r_{\rm max}=50$  and use a relative residual tolerance of $10^{-2}$. 
\change{For the SAA methods, we use the entire training dataset at each iteration.}
For ADAM, we use the final depth of the network and train with a batch size of $32$ and a learning rate of $10^{-3}$.  
For all optimization strategies, we use Tikhonov regularization with $\alpha_1 = \alpha_2 = 10^{-3}$.  
We display the prediction results in \Cref{fig:IndianPinesResults} and plot the convergence histories in~\Cref{fig:IndianPinesConvergence}. 

In \Cref{fig:IndianPinesResults}, we see that GNvpro outperforms training without VarPro in terms of convergence of the accuracy and loss. 
In fact, from the confusion matrix, we see the full Gauss-Newton optimization did not correctly predict any pixels belonging to material $1$, $7$, and $9$.  
These groups contained the fewest number of pixels (\Cref{fig:indianPinesDistribution}) and thus are challenging to classify correctly. 
In its inner loop~\cref{eq:Wargmin}, GNvpro solves for a classification matrix based on all pixels, including those belonging to the smallest classes, resulting in better classification overall. 
We note our results could be further improved by, e.g., taking spatial information into account; see, for instance,~\cite{DBLP:journals/corr/abs-1902-06701}.

In this experiment, GNvpro achieves the best solution, in terms of validation, test accuracy, and computational costs; for example, it uses only $60\%$ the number of work units that we used for ADAM and improves the test accuracy by more than $2\%$.  
The training efficiency can be translated to faster runtimes by exploiting the straightforward parallelism of GNvpro.  
In addition, the overhead of solving for the optimal linear transformation is minimal, taking only $6.4\%$ of the total training time. 
We highlight the improved generalization of VarPro training, since a popular belief is that SAA methods  overfit more easily.
VarPro improves our ability to reduce the loss function, which offers new opportunities to improve generalization by well-crafted direct regularization in contrast to the implicit generalization provided by ADAM.
Through this image segmentation problem with cross-entropy loss, we demonstrate that both GNvpro and L-BFGSvpro converge faster and yield more solutions that generalizes better compared to Gauss-Newton without VarPro and the SGD variant ADAM.

% ============================================================= %
% \section{Stochastic Variant of GNvpro on Image Classification}

% \subsection{Stochastic Variant of GNvpro}
%\change{
%As described in the original optimization problem~\eqref{eq:fullObj}, we can form a sample average approximation with a sufficiently-large subsample $\mathcal{T}\subseteq \mathcal{D}_{\rm train}$.  
%Once we train the network weights on this subsample, we can resample from the training set and repeat~\cite{KleywegtEtAl2006}. 
%Using large enough batches (in the previous experiments, we used the entire training set) and incorporating randomness by resampling reduces the tendency of variable projection to overfit. 
%Additionally, subsampling reduces the memory footprint which is necessary when the entire training dataset cannot be stored, e.g., on a GPU.
%

% }

\subsection{CIFAR-10}
\label{sec:cifar10}

\begin{figure}
    \centering
     \includegraphics[width=\linewidth]{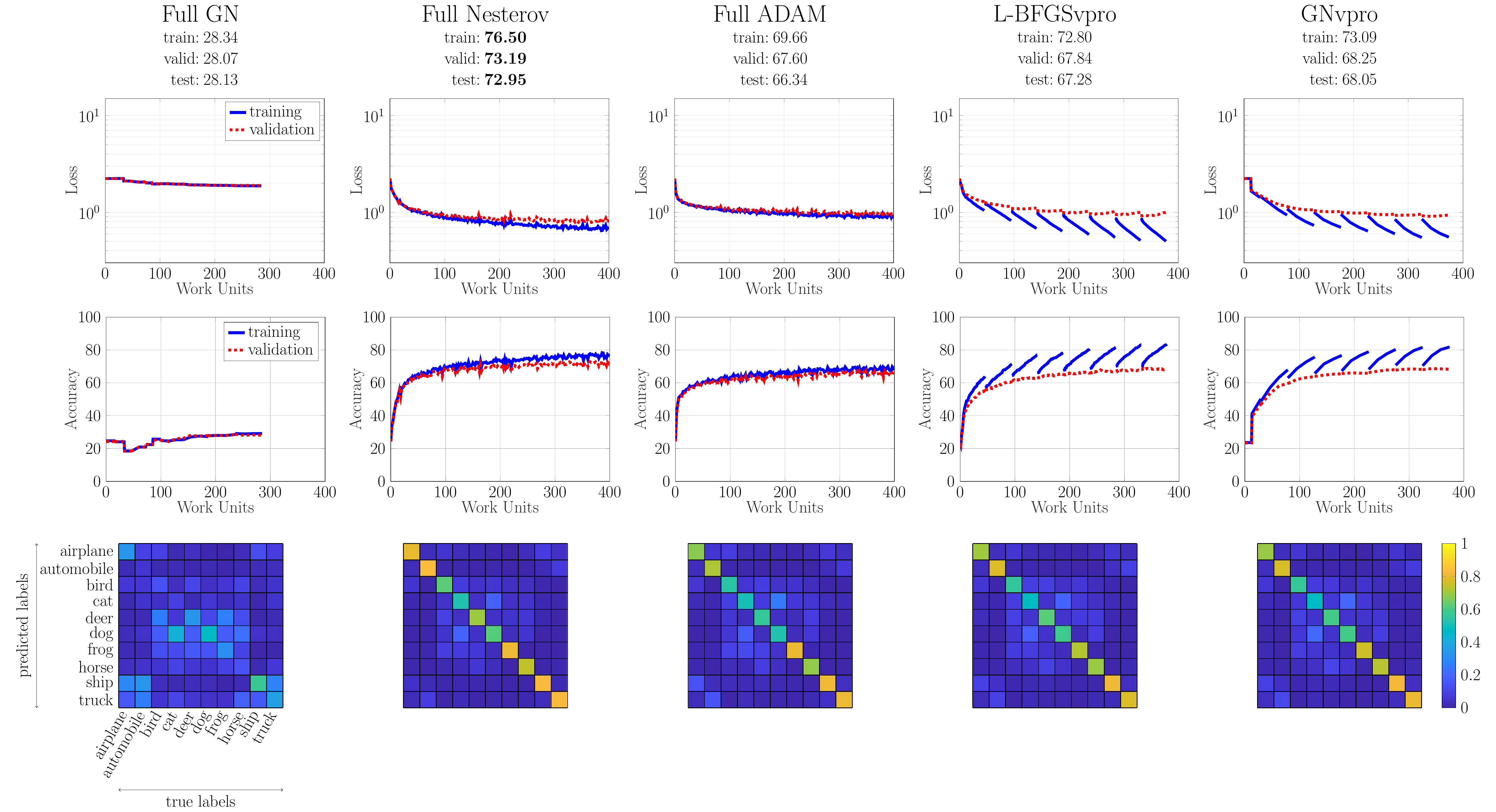}
    \caption{\change{Loss and accuracy convergence plots and confusion matrices of the CIFAR-10 dataset. 
    	Below each method name, we display the accuracy of the trained network on the entire training, validation, and test sets. 
	In this experiment, SGD with Nesterov acceleration achieves the best accuracy across the board. 
	However, both L-BFGSvpro and GNvpro achieve comparable results to ADAM and significantly outperform full Gauss-Newton which does not converge. 
	In the convergence plots for the SAA methods (Full GN, L-BFGSvpro, and GNvpro), we train on four batches of the training data ($10,000$ images per batch) for two epochs. 
	Each solid blue lines (training) indicates the convergence of the loss (top) and accuracy (middle) for one batch. 
	The solid blue lines always start close to the dashed red lines (validation) because the new batch acts like validation data; i.e., data which the network weights has not seen.
	We note that there is a difference between the loss/accuracy of the entire training set and the loss/accuracy of the individual batches. 
	By plotting the training loss for each batch, we observe the rapid decay of the training loss in the VarPro cases.
	Note that in full GN and GNvpro the loss stays constant in the first few work units until a reasonable trust region radius is obtained.
	}}
    \label{fig:cifarConvergenceAccuracy}
\end{figure}

\begin{figure}
    \centering
    \includegraphics[width=\textwidth]{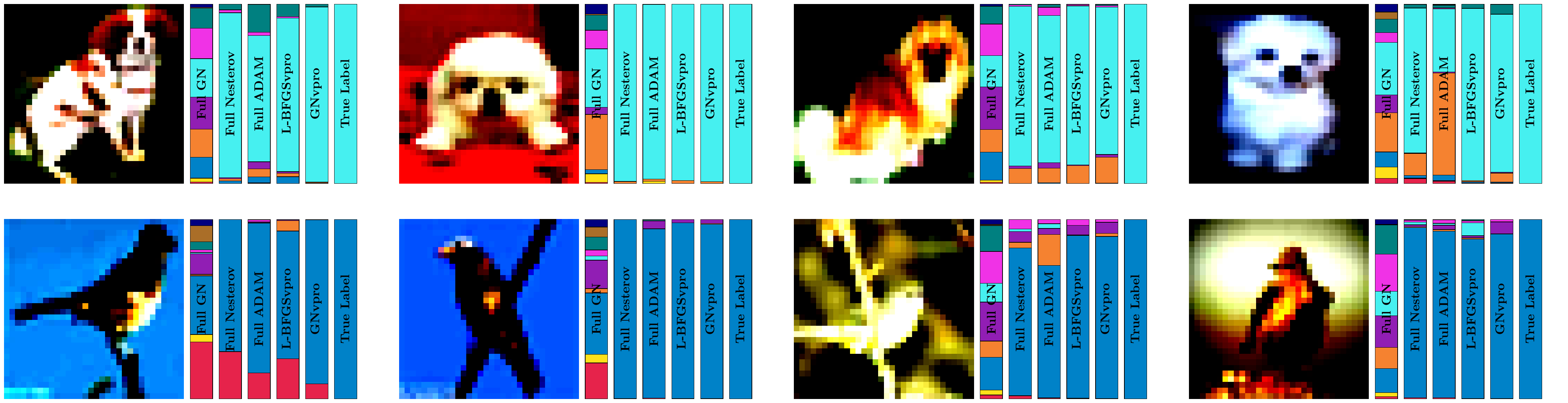}
    \caption{\change{Example image classifications for each optimization method. 
    To the right of each image are size bars corresponding to (left-to-right) Full GN, Full Nesterov, Full ADAM, L-BFGSvpro,  GNvpro, and the true labels, respectively. 
    Each bar is color-coded based on the output probabilities of belonging to each class. 
    The largest color region in each part is the predicted class of each method. 
    The images were chosen such that they were classified correctly by at least four optimization methods.}}
    \label{fig:cifarImages}
\end{figure}

\change{
We use the CIFAR-10 dataset~\cite{Krizhevsky2012;tinyImages} as a commonly-used benchmark problem to demonstrate GNvpro's ability to effectively train convolutional neural networks (CNN) to classify natural images.
The dataset consists of $50,000$ training images  and $10,000$ test images of size $32\times 32 \times 3$ belonging to ten different classes; see~\cref{fig:cifarImages} for examples.
We split the training data into $40,000$ training images and $10,000$ validation images.
Since our main goal is to compare the optimization performance of different algorithms instead of competing with the state-of-the-art results, we consider a relatively shallow CNN architecture similar to the one in~\cite{Malmaud2018}.
Our model consists of two convolutional layers with ReLU activation, each followed by average pooling. 
The network architecture details and hyperparameters chosen are described in detail in~\cref{app:hyperparameterTuning}.
We display the training results in~\cref{fig:cifarConvergenceAccuracy} and a selection of classified images in~\cref{fig:cifarImages}.
% As we show in this experiment, driving down the training loss is a nontrivial task, yet our VarPro methods can do so quickly on the complicated CIFAR-10 dataset\footnote{\change{CIFAR-10 consists of $50,000$ training images  and $10,000$ test images of size $32\times 32 \times 3$ belonging to $10$ different classes; see~\cref{fig:cifarImages} for examples.
% We split the training data into $40,000$ training images and $10,000$ validation images.}}~\cite{Krizhevsky2012;tinyImages}.
}

\change{
We visualize the training performance for the three SAA methods (Gauss-Newton applied to the full problem, L-BFGSvpro, and GNvpro) and two SA methods (ADAM and SGD with Nesterov acceleration) using the blue curves in ~\cref{fig:cifarConvergenceAccuracy}.
As is common in image classification, we measure the performance in terms of classification accuracy and cross entropy loss.
The fact that the full Gauss-Newton scheme did not achieve more than 30\% classification accuracy on the training data underscores the difficulty of solving the training problem with SAA methods. 
In contrast, the training accuracy improves fairly quickly for the SA methods and the SAA methods GNvpro and L-BFGSvpro that use variable projections, which all achieve similar training accuracies ranging from 69.66\% (for ADAM) and 76.50\% (for SGD Nesterov).
In fact, GNvpro and L-BFGSvpro reduce the loss on the current training batch so rapidly that we change the batch after a few iterations; the changes of the dataset can be seen by the jumps in the training loss.
To be precise, we divide the 40,000 training data into four randomly chosen batches each consisting of 10,000 examples and perform two sweeps over the batches.
% , comparing the SAA methods, L-BFGSvpro produces the network with the lowest training and validation loss and highest accuracy and GNvpro achieves the highest test accuracy.
% Training with full Gauss-Newton fails to make progress.
% Comparing to SA methods, Nesterov training produces a network the highest validation and test accuracy, generalizing better for the same number of work units.
% However, ADAM training produces slightly lower validation and test accuracy results than L-BFGSvpro or GNvpro.
}

\change{
The difference between solving the optimization problem in training and accomplishing the learning objective can be seen by comparing the test accuracies plotted as red dotted lines in ~\cref{fig:cifarConvergenceAccuracy}.
In terms of the test accuracies, the best results with 72.95\% are obtained by the SGD method with Nesterov acceleration after some tuning of the learning rate; to see the sensitivity of the results with respect to this parameter, see~\cref{fig:hyperparameterTuningNesterov}.
We note that GNvpro and L-BFGSvpro, since they reduce the training loss more effectively, have  larger generalization gaps than the SGD variants; that is, the difference in training and validation accuracy is larger. 
Despite this gap, both variable projection methods achieve reasonable validation and test accuracies of 67.28\% and 68.05\% for L-BFGSvpro and GNvpro, respectively.
In this example, the performance of GNvpro only depends very mildly on the choice of regularization parameters; see~\cref{fig:hyperparameterTuningNesterov}.
The variable projection methods even slightly outperform ADAM, which achieves a test accuracy of 66.34\%.  
Here, we note the importance of changing the training batches frequently during training in GNvpro and L-BFGSvpro.  
After each change in the training batch, the gap between training loss/accuracy and validation loss/accuracy almost vanishes in ~\cref{fig:cifarConvergenceAccuracy}.  
We observed that without resampling, the training loss continued to decrease while the validation loss stayed roughly constant; i.e., the generalization gap grew.
The generalization gap could possibly be reduced by using larger batches or even the entire training dataset or improved regularization terms.
% , but this is infeasible due to memory and storage considerations.

%   (i.e., driving the training loss down) and the learning problem (i.e., a reasonable fit plus generalization).
% Ideally, we would like these two problems to be synergistic: a good solution satisfies both the optimization problem and the learning problem.
% In practice, these two problems are difficult to solve simultaneously.
% If one solves the optimization problem well, there is no guarantee the solution will generalize beyond the training data (i.e., overfitting).
% If one solves the learning problem well, there is no guarantee the solution will be highly accurate (e.g., the misfit may be large, the gradient of the objective function may be far from zero, etc.).
}

% \TODO{Add overhead cost}

% \change{

% Comparatively, in the stochastic methods, the training and validation loss/accuracy decrease together, with Nesterov achieving better validation and test accuracies than the variable projection methods.
% However, ADAM slightly underperforms GNvpro and L-BFGSvpro.
% The difference in performance between Nesterov and ADAM is one of the challenges to using stochastic optimization methods.
% In our experience, their performance can be difficult to predict and can depend significantly on proper hyperparameter selection, in particular the learning rate.
% In contrast, GNvpro and L-BFGSvpro behave predictably.
% Both algorithms are able to fit the training data quickly and well.
% In addition, we use a trust-region method for GNvpro and a line search satisfying the Wolfe conditions to automatically choose and adapt the step size for our network weight update.

% }

\change{
The takeaways from this experiment are:
	(1) the VarPro methods (GNvpro and L-BFGSvpro) minimize the training loss more efficiently than full optimization methods such as SGD and Gauss-Newton
	(2) resampling the training images every few steps of the SAA method helps reduce the generalization gap
	(3) the SA methods yield a smaller generalization gap, but do not solve the optimization problem efficiently, and 
	(4) in terms of test accuracy, a well-tuned SGD method with Nesterov acceleration performs best on this example followed closely by the VarPro methods that marginally outperform the SA method ADAM.
}

% \TODO{The interpretation of the CIFAR-10 experiment need some work.}

% ============================================================= %
\section{Conclusions}
\label{sec:conclusions}

We present GNvpro, a highly effective Gauss-Newton implementation of variable projection (VarPro) that accelerates the training of deep neural networks (DNNs). 
Our method  extends the reach of VarPro beyond least-squares loss functions to cross-entropy losses to support classification tasks.
As the projection in the non-quadratic case has no closed-form solution, we derive a Newton-Krylov trust region method to efficiently solve the resulting smooth, convex optimization problem. 
Due to our efficient implementation the computational cost of DNN training is dominated by forward and backward propagations through the feature extractor, and the computational costs of the projection step are insignificant in practice.
A key step in the derivation of GNvpro is the computation of matrix-vector products with the Jacobian of the reduced DNN model for which we use implicit differentiation and re-use matrix factorizations from the projection step. 

\change{The key motivation to use VarPro is to exploit the separable structure of most common DNN architectures to obtain training schemes that yield an effective DNN in a few training steps.
This is important since effectively minimizing the training loss is a challenging and necessary first step toward the ultimate goal in deep learning, which is finding a model that generalizes to unseen data. 
In other words, a model that performs poorly on the training data should not be expected to perform better on the test data. 
Our experiments demonstrate that GNvpro and L-BFGSvpro are able to minimize the training loss more efficiently than methods that do not exploit the separable structure of the problem, e.g.,  Gauss-Newton and stochastic gradient methods applied to the full problem.
% Minimizing the training loss with SAA methods is notoriously challenging, as is best illustrated in the CIFAR-10 example (\Cref{sec:cifar10}) where the Gauss-Newton scheme is unable to reduce the training loss.
In addition to the optimization scheme, the ability to generalize also depends on other factors such as the choice of network architecture and regularization strategy and properties of the training data.
While GNvpro achieves a higher test accuracy than the other methods in the surrogate modeling and image segmentation problems, it was slightly inferior to the best-tuned SGD method for the CIFAR-10 image classification problem. 
Possible causes for this could include the relatively small sample size compared to the high dimension and heterogeneity of the training data. 
Note that in this dataset, the images belonging to a given class can be considerably different in their contrast and appearance.
This is different in the hyperspectral imaging example where the measurement device is engineered such that pixels belonging to the same class will have similar spectra.}

The mechanics of VarPro are similar to those of block coordinate descent approaches (e.g., the recent DNN training algorithms in~\cite{patel2020block,cyr2020robust}). 
Added challenges in VarPro are the higher accuracy needed for the inner problem and the slightly more involved derivations of Jacobians for the outer problem when using Gauss-Newton-type approaches.
A key contribution of our paper is the derivation of efficient algorithms and their implementations that limit the computational costs of those parts, rendering the computational costs of GNvpro comparable to block coordinate descent. 
A crucial advantage added by GNvpro is its ability to  account for the expected coupling between $\bfW$ and $\bftheta$ and therefore may be more efficient (for some evidence, see the imaging examples in~\cite{ChungEtAl2006}).

To demonstrate the applicability of GNvpro, our numerical experiments vary the DNN architecture to accomplish several learning tasks, \change{namely surrogate modeling, image segmentation, and image classification}.
We provide numerical experiments for multilayer perceptron , a shallow convolutional network,  and Neural ODEs, different loss functions (linear regression, logistic regression, multinomial regression), and optimization methods (Gauss-Newton, BFGS, ADAM).
Across all of these experiments, GNvpro trains the networks more efficiently, i.e., it achieves a lower training loss for a given number of work units.  
\change{Solving the training problem is non-trivial as can be seen by the poor performance by the full Gauss-Newton scheme.}
Furthermore, \change{in all but the classification experiment, GNvpro also achieved} high validation accuracy indicates that the trained DNNs also generalize well.
We note that in addition to fewer work units, the straightforward parallelism offered by SAA methods can dramatically reduce the computational time of DNN training.

Our numerical experiments suggests that GNvpro has a few traits that make it particularly useful for PDE surrogate modeling. 
Most importantly, VarPro improves the accuracy of the learned parameter-to-observable map by an order of magnitude compared to both SAA and SGD methods that do not exploit the structure of the architecture. 
\change{In our examples, the data is obtained from an accurate PDE solver and is not corrupted by noise. This assumption is reasonable in practical applications, e.g., when using computationally-efficient surrogate models of complex physical systems. }
VarPro may also provide new avenues for solving other learning problems for which existing methods fail to reduce the loss function sufficiently; for example, scientific applications such as solving PDEs using neural networks. 
\change{In the presence of inaccurate or noisy data,  SAA methods like GNvpro can be combined with well-established computational tools such as Generalized Cross Validation and efficient hybrid schemes (using direct and iterative regularization) and potentially automatically tune regularization parameters~\cite{ChungNagyOLeary:GCV2008, Bjork:ImplicitShift1994, Hansen:DiscreteInverseProblems2010}.
Such tools cannot be applied in the ADAM framework easily. 
Therefore, our results provide the opportunity for another exciting future research direction.}

Our primary reason to use an SAA method is to avoid overfitting when solving the inner optimization problem \cref{eq:Wargmin}.
As SAA methods generally use larger batch sizes compared to stochastic approximation methods such as SGD, the risk is reduced. 
Even though the SAA methods perform well in our examples, exploring the use of VarPro in SGD-type methods is an interesting open question.

For future work, we will investigate the benefits of GNvpro across an even wider range of learning problems. 
To this end, we make our prototype implementation freely available as part of the {\tt Meganet.m} package, a public {\sc Matlab} neural network repository also used in~\cite{HaberRuthotto2017, RuthottoHaber2019}.
To further its reach, we also plan to implement GNvpro in other machine learning frameworks such as TensorFlow and pyTorch.

\section*{Acknowledgments}
We thank Eldad Haber for introducing us to Variable Projection and many fruitful discussions. 
% \change{We also thank our reviewers for their helpful comments.}
Any subjective views or opinions that might be expressed in the paper do not necessarily represent the views of the U.S. Department of Energy or the United States Government. Sandia National Laboratories is a multimission laboratory managed and operated by National Technology and Engineering Solutions of Sandia LLC, a wholly owned subsidiary of Honeywell International, Inc., for the U.S. Department of Energy’s National Nuclear Security Administration under contract DE- NA-0003525. SAND2020-8481 J.

%{\color{red} \bf IN PROGRESS: Add more references. Also delete duplicates please.}

\bibliographystyle{siamplain}
\bibliography{main.bib}

\appendix

% =========================================================================== %
\section{The Importance of Solving the Inner Optimization Problem Well}	
\label{app:innerProblem}
\change{
    The accuracy of the gradient $\nabla_{\bftheta}\Phi_{\rm red}(\bftheta)$, and thus the effectiveness of the DNN training, depends crucially on how well we solve the inner optimization problem, specifically we need the norm of $\nabla_{\bfw} \Phi(\bfW(\weights),\weights)$ to be small.
	The norm of this gradient will be affected by the accuracy of the inner optimization problem (tolerance) and the dimension of the Krylov subspace when solving for $\bfW$.		
	
	We note that in~\eqref{eq:varproGrad}, there is a multiplication of $J_{\bftheta} \bfw(\bftheta)^\top\nabla_{\bfw} \Phi(\bfW(\weights),\weights)$.
	This Jacobian operator is derived in~\eqref{eq:implicitSolve} as the product of two matrices.  
	The first matrix is the inverse Hessian $\nabla_{\bfw}^2 \Phi(\bfW(\weights),\weights)^{-1}$ for which the condition number can be controlled by the regularization term on $\bfW$. 
	The second matrix depends on the Jacobian of the network, which is hard to define explicitly, but has not been problematic in our experiments. 
	In practice, we have found that if $\nabla_{\bfw} \Phi(\bfW(\weights),\weights)$ is small, then $J_{\bftheta} \bfw(\bftheta)^\top\nabla_{\bfw} \Phi(\bfW(\weights),\weights)$ will be small as well.
	
	% \todo{Hence, the multiplication of $J_{\bftheta}\bfw(\bftheta)$  by a vector close to zero should remain negligible.}

        	To assess the importance of solving the inner optimization problem well, we perform the following experiment.
	We compute the gradient of the reduced objective function for various choices of tolerance and Krylov space rank when solving the inner optimization problem. 
	The data, network, and hyperparameter selection are not the focus of this experiment, and are fixed\footnote{\color{black}We use the \texttt{peaks} classification data (see, e.g.,~\cite{HaberRuthotto2017}) and a ResNet with a width of $4$ and a depth of $8$ corresponding to a final time of $5$.  
	We use regularization parameters $\alpha_1=5\cdot 10^{-6}$ on $\bftheta$ and $\alpha_2 = 10^{-8}$ on $\bfW$.}.
	
    We measure the error of the first-order Taylor approximation
        \begin{align*}
            \text{error} = |\Phi_{\rm red}(\bftheta) + h\nabla\Phi_{\rm red}(\bftheta)^\top\delta\bftheta - \Phi_{\rm red}(\bftheta + h\delta\bftheta)| 
        \end{align*}
        where $\delta\bftheta$ is a random perturbation of the weights $\bftheta$. 
        If the gradient $\nabla\Phi_{\rm red}(\bftheta)$ is sufficiently accurate, we should see the error converge like $O(h^2)$ as $h \to 0$.  
        We expect the gradient to be accurate when the inner optimization is solved to a sufficient tolerance and when the rank of the Krylov space is large enough to render $\|\nabla_{\bfw} \Phi(\bfW(\bftheta), \bftheta)\|$ sufficiently small.
	We observe this behavior in~\cref{fig:subproblemAccuracy}; a sufficiently small $\|\nabla_{\bfw} \Phi(\bfW(\bftheta), \bftheta)\|$ and a reasonable Krylov dimension for the inner optimization problem ensure the gradient of the reduced objective function is accurate.
	% Additionally, we display he spectrum of $\nabla_{\bfw}^2\Phi(\bfW(\bftheta),\bftheta)$ in~\cref{fig:hessWEigenvalues} to support the need to solve the inner optimization problem well. 

    \begin{figure}
        \centering
        \includegraphics[width=\textwidth]{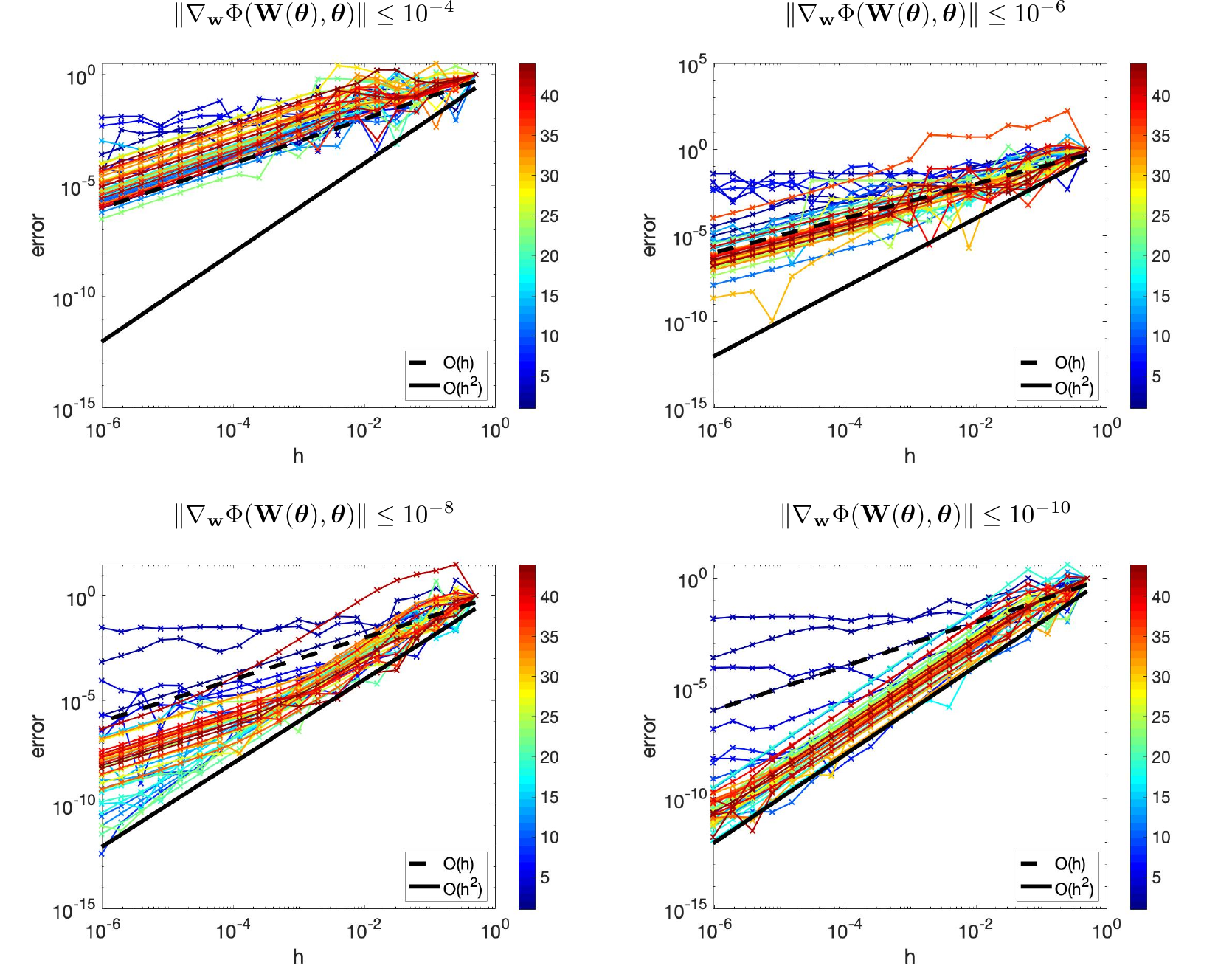}
        \caption{\color{black} Accuracy of $\nabla \Phi_{\rm red}(\bftheta)$ for various choices of Krylov rank when solving the inner optimization problem. 
       	Each color represents a given maximum Krylov rank, $r_{\rm max}$.
        Each image has a different tolerance on the inner optimization problem.
        Ideally, all of the curves should track with the solid black line representing $O(h^2)$ convergence. 
        This indicates our gradient $\nabla \Phi_{\rm red}(\bftheta)$ is correct and we capture the linear approximation perfectly.
        We observe that once the tolerance of the inner optimization problem is small enough (bottom right),  the gradient $\nabla \Phi_{\rm red}(\bftheta)$ is accurate for a reasonable Krylov rank (conservatively, $r_{\rm max}\gtrsim 15$). 
        The tolerance is influences the accuracy most significantly.
        }
        \label{fig:subproblemAccuracy}
    \end{figure}
}

% =========================================================================== %
\section{Derivation of the Jacobian for Cross-Entropy}	
\label{app:jacobian}
\change{
	To provide more derivation details, let $\bfz(\bftheta) = F(\bfy,\bftheta)$ be the output features for one sample. 
	The gradient of the full objective function with respect to $\bfW$ can be written in several equivalent variations, specifically
    	\begin{align*}
	\nabla_{\bfw}\Phi(\bfW,\bfc) 
		&=\text{vec}(\nabla L(\bfW\bfz(\bftheta),\bfc) \bfz(\bftheta)^\top)\\
		&\Leftrightarrow (\bfz(\bftheta) \otimes \bfI_{\nTargets}) \nabla L(\bfW\bfz(\bftheta),\bfc)  \\
		&\Leftrightarrow (\bfI_{\nFeatOut} \otimes \nabla L(\bfW\bfz(\bftheta),\bfc)) \bfz(\bftheta)
	\end{align*}
	where $\nabla L$ is the gradient of the loss with respect to the first argument and $\otimes$ is the Kronecker product.
	We omit the regularization term for simplicity.
	
	%\medskip
	
	For any loss function, the Jacobian of the above gradient with respect to $\bftheta$ is
	\begin{align*}
	J_{\bftheta}\nabla_{\bfw}\Phi(\bfW,\bfc) 
		&=(\bfz(\bftheta) \otimes \bfI_{\nTargets}) \nabla^2 L(\bfW\bfz(\bftheta),\bfc)\bfW\bfJ_{\bftheta}\bfz(\bftheta)
			+(\bfI_{\nFeatOut} \otimes \nabla L(\bfW\bfz(\bftheta),\bfc))\bfJ_{\bftheta}\bfz(\bftheta).
	\end{align*}
	In the case of cross-entropy, the gradient of the loss with respect to the first argument has the form
		\begin{align*}
		\nabla L(\bfW\bfz(\bftheta),\bfc) 
			&=\frac{1}{|\mathcal{T}|}\left(-\bfc + \frac{\exp(\bfW\bfz(\bftheta))}{\bfe^\top \exp(\bfW\bfz(\bftheta))}\right)
		\end{align*}
	where $\bfe\in \R^{\nTargets}$ is the vector of all ones. 
	The Hessian of the loss with respect to the first argument for one sample can be expressed as 
		\begin{align*}
		\nabla^2 L(\bfW\bfz(\bftheta),\bfc) 
			= \frac{1}{|\mathcal{T}|}\left[
			\diag\left(\frac{\exp(\bfg(\bftheta))}{\bfe^\top \exp(\bfg(\bftheta))}\right)-\frac{\exp(\bfg(\bftheta))\exp(\bfg(\bftheta))^\top}{(\bfe^\top \exp(\bfg(\bftheta)))^2}
			\right]
		\end{align*}
	where $\bfg(\bftheta) = \bfW\bfz(\bftheta)$ for simplicity. 

}

% =========================================================================== %
\section{Data Generation for Surrogate Modeling Examples} 
\label{appendix}

To supplement \Cref{sub:surrogateModeling}, we provide details about the PDE models. 

% =========================================================================== %
\paragraph{Convection Diffusion Reaction (CDR)}

\begin{figure}
    \centering
    \includegraphics[width=.32\textwidth]{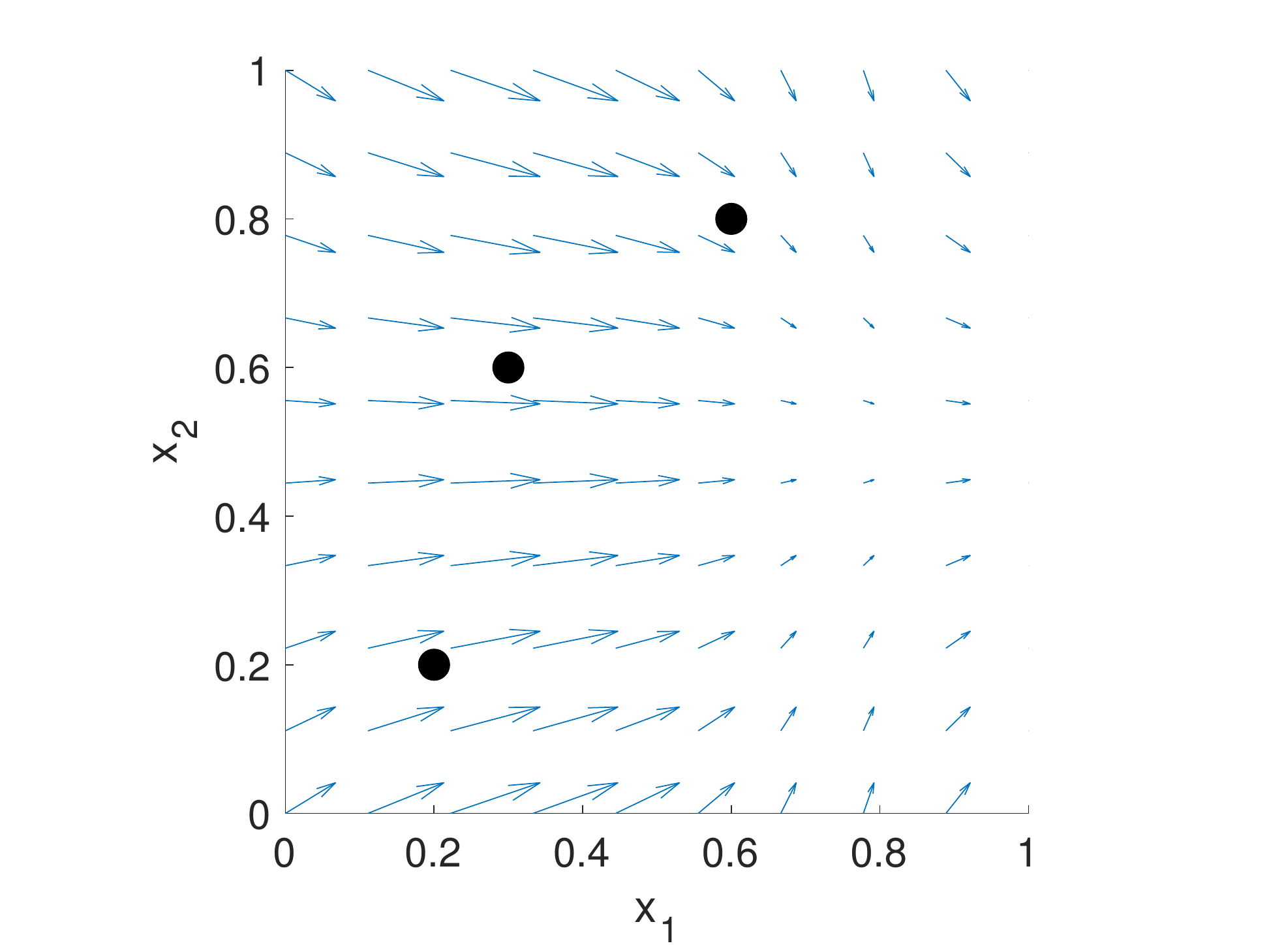}
    \includegraphics[width=.32\textwidth]{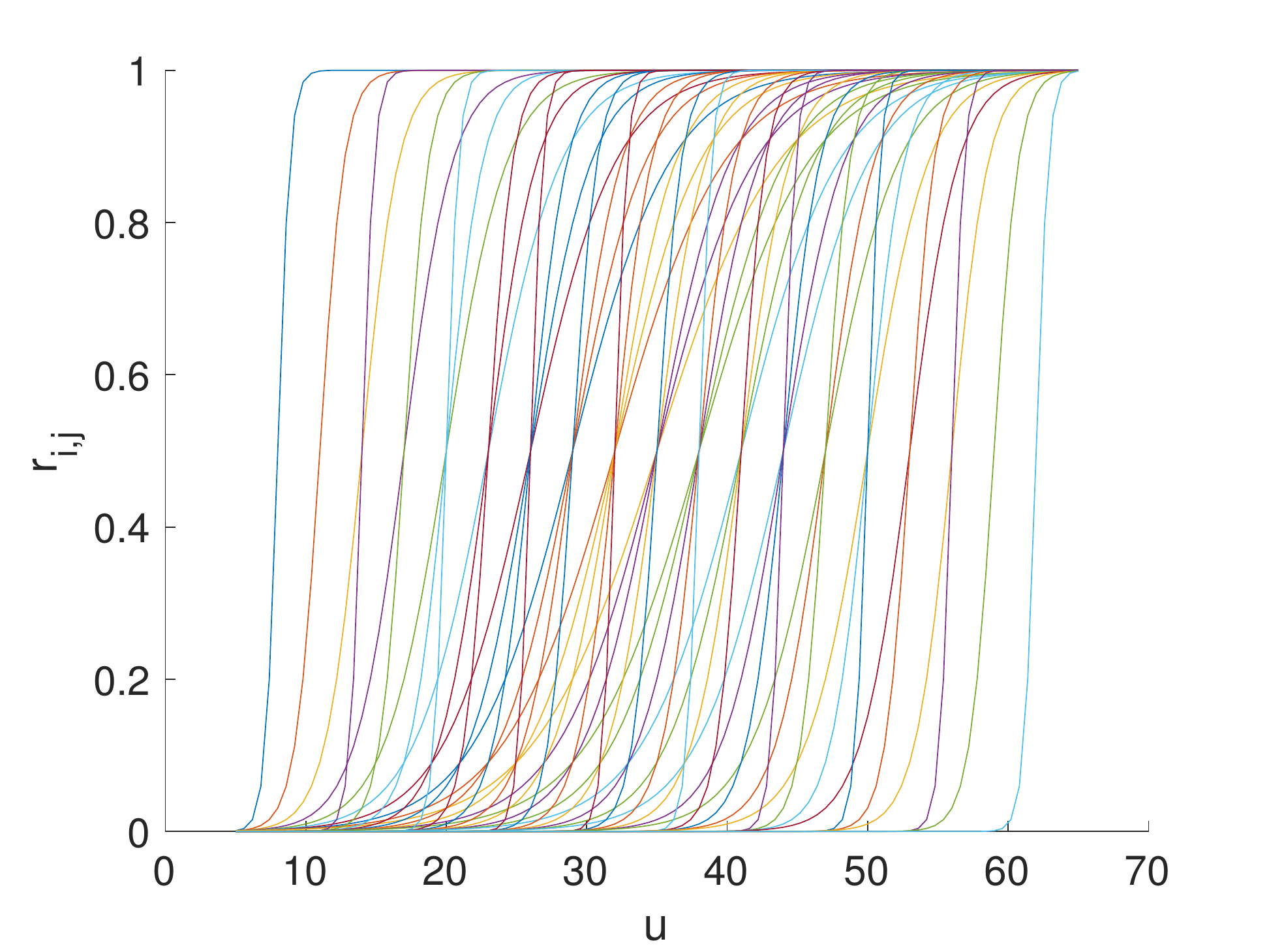}
    \includegraphics[width=.32\textwidth]{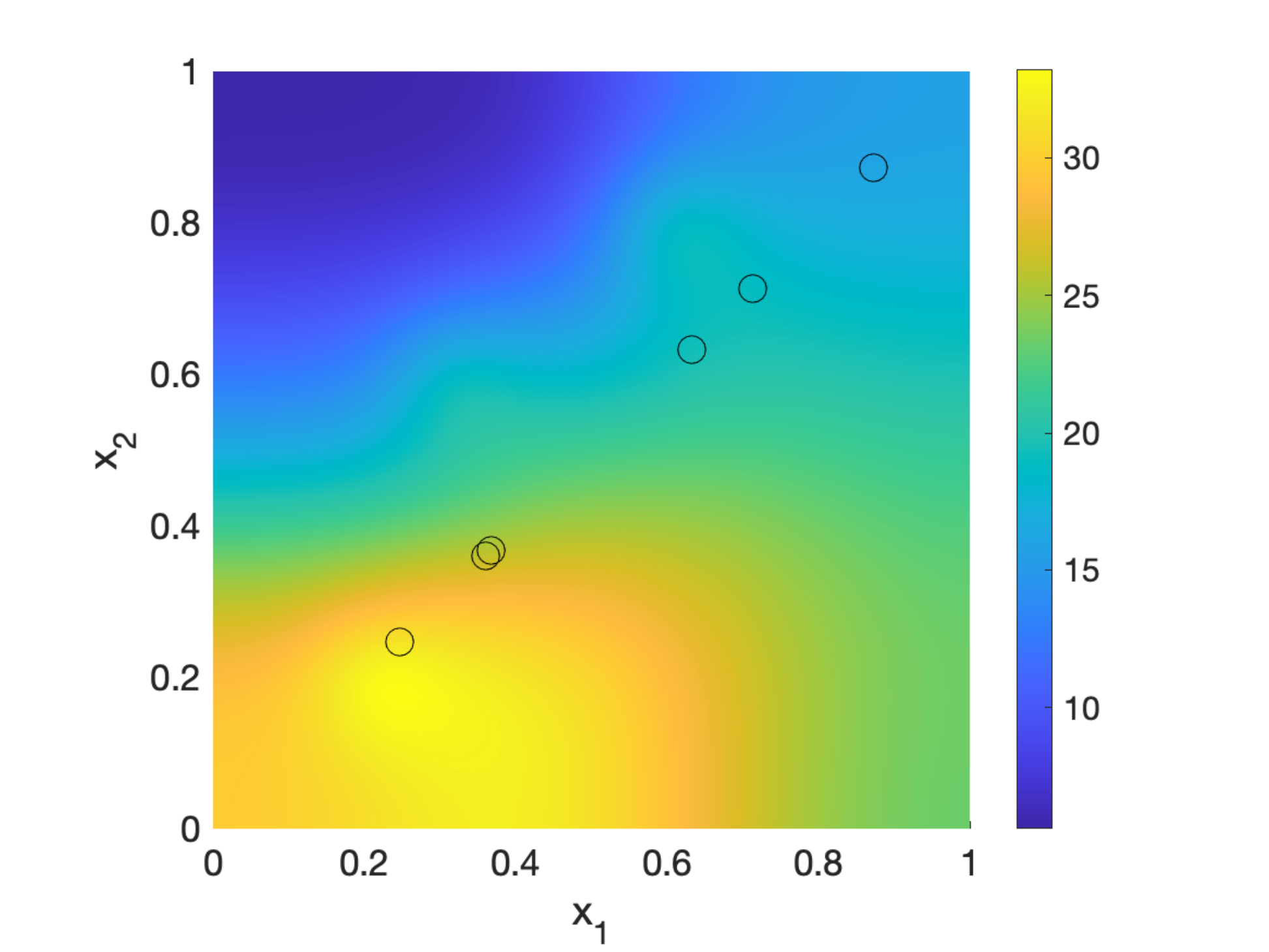}
    \caption{Visualizing the generation for the Convection Diffusion Reaction dataset.  Left: velocity field $\mathbf v$ with direction denoted by the arrow heads and magnitude by the arrow size, and the Gaussian source locations indicated by dots; center: plot of each reaction functions $\{r_{i,j}\}$; right: time $t=0.5$ snapshot of a state realization with the observation locations indicated by circles.}
    \label{fig:CDR_PDE_Plots}
\end{figure}

We consider the CDR equation defined on the time interval $[0,0.5]$ and spatial domain $\Omega=(0,1)^2$ with a zero initial condition and homogeneous Neumann conditions on the boundary $\partial \Omega = [0,1]^2 \setminus \Omega$, i.e.
	\begin{align*}\label{eq:cdr_model}
	& \frac{\partial u}{\partial t}  = \nabla \cdot (D\nabla u) - \bfv \cdot \nabla u + f + \bfy^T \mathbf r(u) \qquad & \text{on } \Omega \times (0,0.5] \\
	& D \nabla u \cdot \mathbf n = 0 \qquad & \text{on } \partial \Omega \times (0,0.5] \\
	& u = 0 \qquad & \text{on } \Omega \cup \partial \Omega \times \{0\}
	\end{align*}
where $u:[0,1]^2 \times [0,0.5] \to \R$ is the state variable (e.g., concentration), $D=0.5$ is the diffusion coefficient, $\bfv:(0,1)^2 \to \R^2$ is the (stationary) velocity field given by
\begin{eqnarray*}
\bfv(x_1,x_2) = \left( 4.5 \left(1+0.8 \sin(2\pi x_1) \right), 5.5 \left(0.5-x_2 \right) \right),
\end{eqnarray*}
and $f:[0,1]^2 \times [0,0.5] \to \R$ is the source term given by 
\begin{align*}
f(x_1,x_2,t) =& \hspace{.25 cm} 1000\left( 1.0+0.3 \cos(2 \pi t) \right) \exp(-200\left( (x_1-0.2)^2 + (x_2-0.2)^2 \right) ) \\
& +  800\left( 1.2+0.4 \sin(4 \pi t) \right) \exp(-200\left( (x_1-0.6)^2 + (x_2-0.8)^2 \right) ) \\
& + 900\left( 0.9+0.2 \sin(6 \pi t) \right) \exp(-200\left( (x_1-0.3)^2 + (x_2-0.6)^2 \right) ).
\end{align*}
The velocity field $\mathbf v$ and centers of the Gaussian source locations are shown in the left panel of Figure~\ref{fig:CDR_PDE_Plots}. The reaction term $\bfy^T \mathbf r(u)$ (a nonlinear function of the state $u$) is the inner product of the parameter vector $\bfy \in \mathbb R^{55}$ and 
\begin{eqnarray*}
\mathbf r(u) = \overline{r} \left( r_{1,2}(u),r_{1,3}(u),\dots,r_{1,11(u)},r_{2,3}(u),r_{2,4}(u),\dots,r_{2,11}(u),r_{3,4}(u),\dots,r_{10,11}(u) \right)^\top
\end{eqnarray*}
which combines the reaction functions
\begin{eqnarray*}
r_{i,j}(u) =  \left(\frac{C\exp{\left(\ln\left({C^{-2}}\right) \frac{u-u_i}{u_j-u_i}\right)}}{1+C\exp{\left(\ln\left({C^{-2}}\right) \frac{u-u_i}{u_j-u_i}\right)}} \right),
\end{eqnarray*}
where $C = \frac{10^{-3}}{1-10^{-3}}$ and $u_i = 5 + 6(i-1)$, $i=1,2,\dots,11$,
and the spatially heterogeneous scaling field
\begin{align*}
\overline{r}(x_1,x_2) & = 500 \left( 0.8 + 0.2\sin(2\pi x_1)\cos(2\pi x_2) + x_2 \right).
\end{align*}

The reaction functions $\{r_{i,j}\}$ are a set of $55$ sigmoid functions which activate (begin taking values $\gg0$) and saturate (approach their vertical asymptote 1) for different values of $u$, i.e. they model reactions which occur over various concentration ranges. The center panel of Figure~\ref{fig:CDR_PDE_Plots} shows each $r_{i,j}$ plotted as a function of $u$. 

The state $u$ is initially zero, but increases through the time dependent forcing $f$, is advected toward $(1.0,0.5)$ by the velocity field $\mathbf v$, and undergoes diffusion and spatially heterogeneous reaction.  The right panel of Figure~\ref{fig:CDR_PDE_Plots} displays a typical realization of the state $u$ at the final time $t=0.5$.

We solve the PDE using linear finite elements on a rectangular mesh containing 151 nodes in each spatial dimension, and 51 steps using backward Euler on a uniform time grid. 
%It takes on the order of five minutes, using a single processor of a Linux workstation, to solve the PDE for a given $\bfy$.

The reaction function coefficients (the input data) $\bfy \in \mathbb R^{55}$ are generated by sampling each component of $\bfy$ independently from a uniform distribution on $[0,1]$ and subsequently normalizing by the sum of components. The target data $\bfc$ is generated by evaluating the solution $u(x_1,x_2,t)$ at six spatial locations (indicated by the circles overlaying the plot in the right panel of Figure~\ref{fig:CDR_PDE_Plots}) for each of the final twelve time steps, i.e. at times $t=0.39+0.01k$, $k=0,1,\dots,11$.

% =========================================================================== %
\paragraph{Direction Current Resistivity (DCR)}

\begin{figure}[t]
	\centering
	\includegraphics[width=1\textwidth]{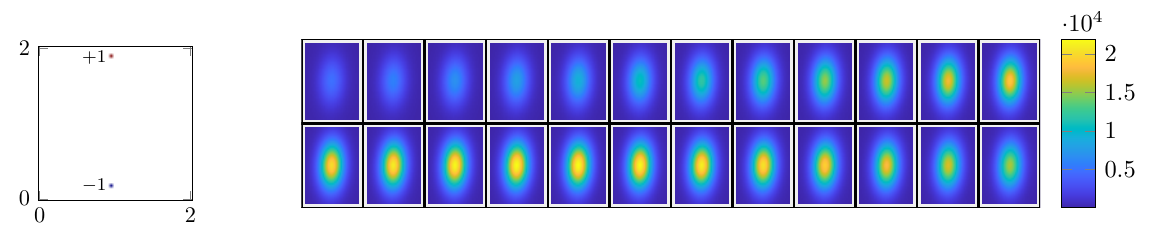}
	\caption{Visualizing the source and conductivity region for the DC Resistivity dataset.  
	Left: The dipole source $q$ in the $x_1-x_2$ plane at the surface $x_3=1$.  
	The dipole source is $-1$ at the coordinate $(1,\frac{1}{6})$ and is $+1$ at the coordinate $(1, \frac{11}{6})$, $q(x_1,x_2,x_3)=0$ for $x_3 < 1$;
	right: Data generation of the DC Resistivity data set. We visualize the conductivity, $m$, to detect.  Each image is a slice parallel to the $x_1$-$x_2$-plane.  The top left is the surface, and left-to-right, top-to-bottom progresses down in depth.}
	\label{fig:dipoleConductivity}
	\end{figure}

The PDE modeling the electric potential in the DCR example is given by Poisson's equation on the spatial domain $\Omega = (0,2)\times(0,2)\times(0,1)$ with homogenous Neumann conditions on $\partial \Omega = [0,2] \times [0,2] \times [0,1] \setminus \Omega$, i.e.
    \begin{align*}\label{eq:pde continuous}
        &-\nabla \cdot \left(m(\cdot,\bfy) \nabla u\right) = q \qquad & \text{on } \Omega \\
        & \nabla u \cdot \mathbf n = 0 \qquad & \text{on } \partial \Omega
    \end{align*}
where $u: \Omega \to \R$ is the state variable (e.g., the electric potential) and $q:\Omega\to \R$ is a source (e.g., a dipole) visualized in \Cref{fig:dipoleConductivity}.

	The model or control variable $m: \Omega \times \R^{\nFeatIn} \to \R_+$ (e.g., conductivity), parameterized by $\bfy\in \R^3$, is given by
    \begin{align*}\label{eq:sigma}
        m(\mathbf x;\bfy) = \exp\left(10\exp\left(-\frac{1}{y_2^2}\left(\left\|\diag(\sqrt{2},1)\bfQ(y_3)\begin{pmatrix} x_1-1 \\ x_2-1\end{pmatrix}\right\|^2+(x_3-y_1)^2\right)\right)\right),
    \end{align*}
where $\mathbf x = (x_1,x_2,x_3)$ and $\bfQ(y_3)$ is the $2\times 2$ rotation matrix parameterized by the angle $y_3$.	
We visualize one instance of the ellipsoidal conductivity region to detect in \Cref{fig:dipoleConductivity}. 
%Note that we could also allow $\bfy$ to correspond to a discretization of a spatially-dependent variable (i.e., infinite-dimensional) conductivity field. 

We solve the PDE using a Finite Volume approach~\cite{Haber2015} with a rectangular mesh containing $48$, $48$, and $24$ cells in each dimension, respectively.
% Our discretized PDE operator is a $60025 \times 60025$ sparse ($99\%$ zeros), symmetric positive definite matrix, which we invert to solve for the function $u$. 

The model parameters (the input data) $\bfy\in \R^3$ are generated by sampling from a uniform distribution on $[0,1]\times [0.25, 3] \times [0,\pi]$.  
The target data $\bfc\in \R^{882}$ is generated by measuring the differences in the electric potential $u$ at points in a uniform $21 \times 21$ rectangular grid in the $x_1$-$x_2$ plane at the surface ($x_3=1$). Collecting the differences in the $x_1$ and $x_2$ directions at each of these 441 points gives $\bfc \in \R^{882}$. Because the data was generated with a large dipole source, we subtract the mean from all the samples so $\bfc$ is the distinguishing features rather than the dominant source.

% =========================================================================== %
\section{CIFAR-10 Experimental Setup}
\label{app:hyperparameterTuning}
\change{The convolutional neural network architecture is summarized in~\cref{tab:cifarCNN}. 
We end with a final affine layer to fit the $64$ output features to the $10$ classes. 
We initialize the weights of the nonlinear feature extractor, $\bftheta$, using the Kaiming uniform initialization~\cite{he2015delving} and, for the non-VarPro cases, initialize the weights of the linear classifier $\bfW$ using variable projection on the full training dataset. }

\change{
The run-time of the inner optimization for GNvpro is about $14.2\%$ of the entire training time.  
We note that for the CIFAR-10 experiment, there is a GPU-CPU transfer in our prototype implementation which can be improved.  
	We train the network weights $\bftheta$ on the GPU, but solve for $\bfW$ on the CPU, hence the slightly increased percentage of run-time for the inner optimization problem.  
	The overhead is still small given that we can solve the optimization problem (i.e., fit the training data) so quickly.
	We further mention the network in our CIFAR-10 experiments is very shallow.  
	For deeper networks, the run-time to solve for $\bfW$ will be a smaller percentage of the total training run-time.
}

\begin{table}
\centering
\scriptsize
\caption{\change{CIFAR-10 Convolutional Neural Network architecture and learnable weights.  
 $\bfW$ corresponds to the weights of the final affine layer. For the reduced problem, $\bfW$ is optimized explicitly, and hence does not count as a learnable parameter.}}
\label{tab:cifarCNN}
\change{
\begin{tabular}{|l|c|c|c|c|}
\hline
Layer Type &Description & \# of Output Features & \# of Weights\\
\hline\hline
Convolution + ReLU & $32$, $5\times 5 \times 3$ filters, stride $1$ & $32\times 32\times 32$ & $2400$\\
\hline
Average Pooling & $2\times 2$ pool, stride $2$& $16\times 16\times 32$ & $\--$\\
\hline
Convolution + ReLU & $64$, $5\times 5 \times 32$ filters,  stride $1$  & $16\times 16\times 64$ & $51200$\\
\hline
Average Pooling & $16\times 16$ pool, stride $16$& $1\times 1\times 64$ & $\--$\\
\hline
Affine Layer & $10\times 64$ matrix + $10\times 1$ vector & $10\times 1$  & $650$\\
\hline\hline
\multicolumn{3}{|l|}{Total} & $53600$ $(+650)$\\
\hline
\end{tabular}
}
\end{table}

\begin{figure}
\centering
\begin{tabular}{m{0.025\textwidth}m{0.9\textwidth}}
\rotatebox{90}{Loss} & \includegraphics[width=0.9\textwidth]{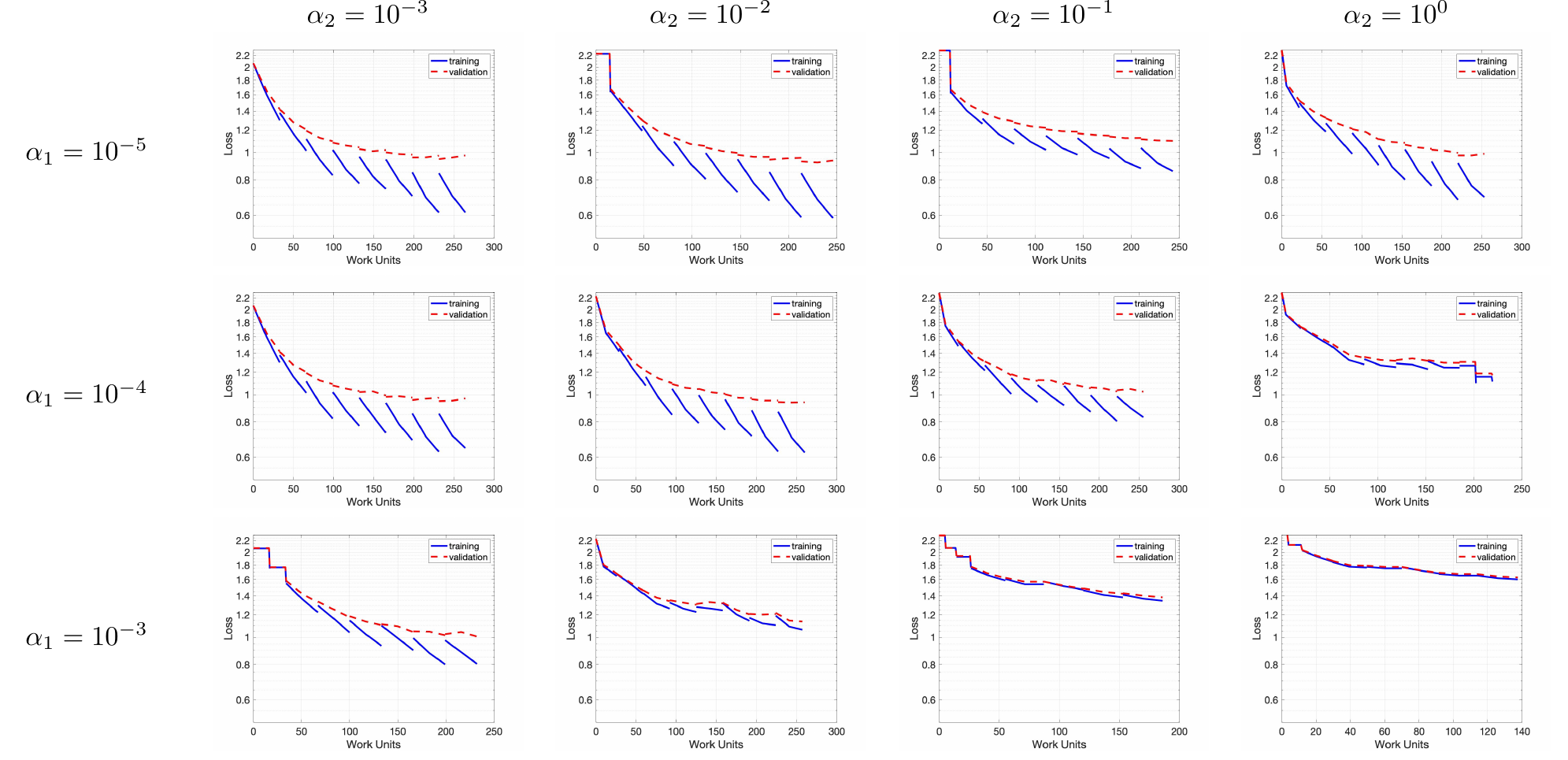}\\[0.5em]
\hline\\[-0.5em]
\rotatebox{90}{Accuracy}  & \includegraphics[width=0.9\textwidth]{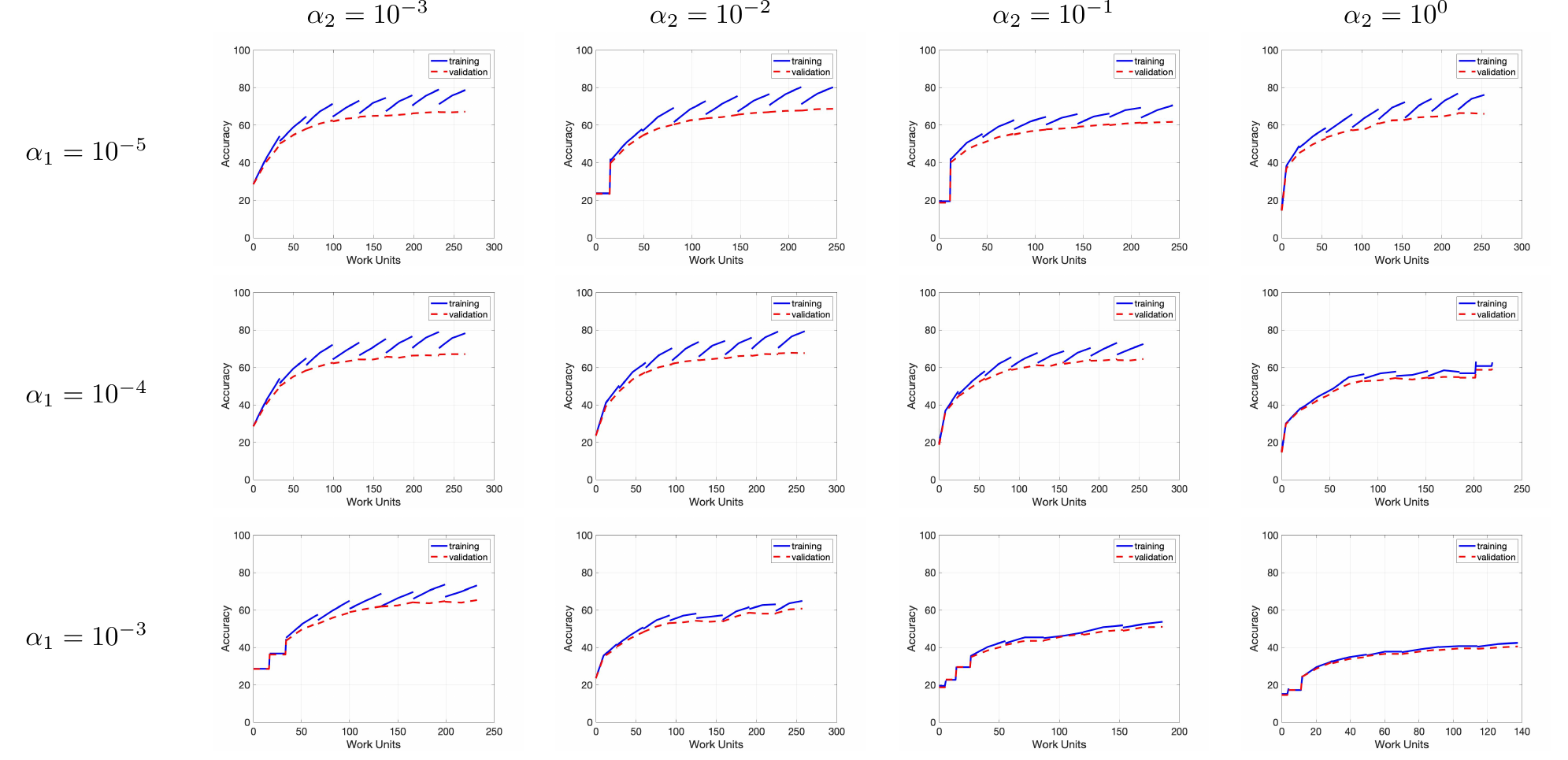}
\end{tabular}
\caption{\change{GNvpro regularization parameter tuning CIFAR-10 experiment.  
Regularization parameters were chosen to give the best validation accuracy while providing fast convergence of the training loss without observing significant semiconvergence behavior on the validation loss.. 
We chose $\alpha_1 = 10^{-4}$ and $\alpha_2 = 10^{-2}$. 
Other candidates were possible, but would not have significantly impacted the presented results.}}
\label{fig:hyperparameterTuningGNvpro}
\end{figure}

\begin{figure}
\centering
\includegraphics[width=0.9\textwidth]{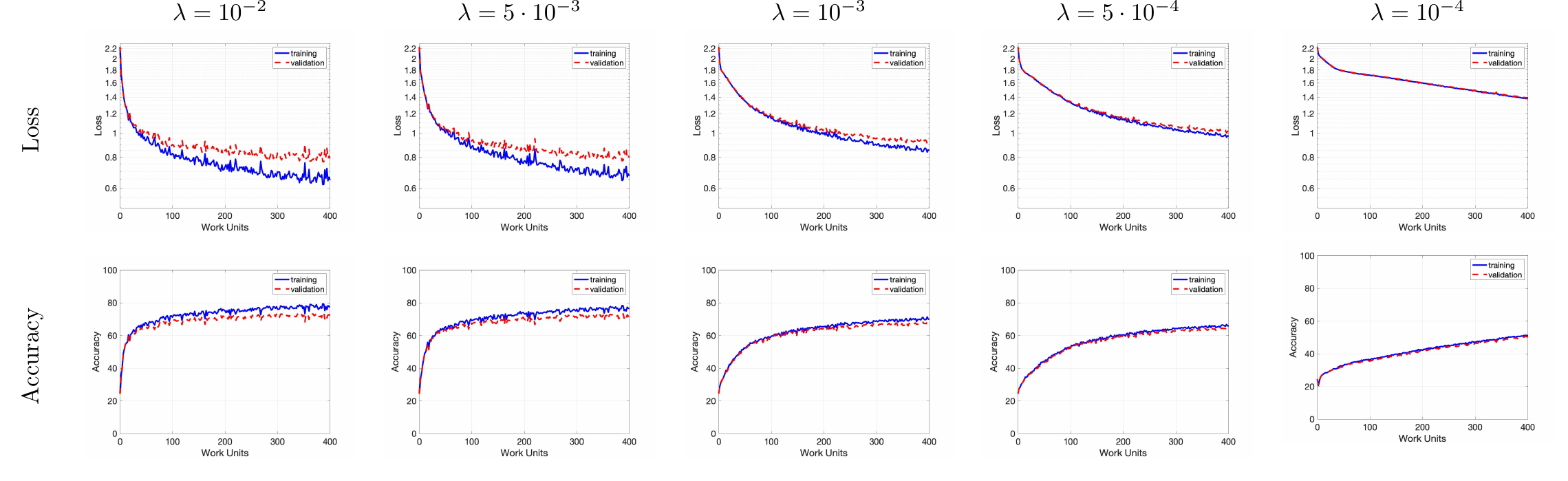}
\caption{\change{Nesterov learning rate tuning for the CIFAR-10 experiment.  
The learning rate was chosen to give the best validation accuracy while maintaining a small generalization gap.  
There was a noticeable accuracy advantage of using $\lambda=5\cdot 10^{-3}$ over the traditionally-recommended $10^{-3}$ in this experiment.}}
\label{fig:hyperparameterTuningNesterov}
\end{figure}

\change{
We obtained these hyperparameters by training the network with various combinations of parameter choices and selecting the best performance.
In the SAA method, we varied the rank of the Krylov space, the number of trust-region iterations, the batch size, the number of epochs, and the regularization parameters and chose the options that gave GNvpro a good validation accuracy (see~\cref{fig:hyperparameterTuningGNvpro} for examples).
We used these same parameters for full GN and L-BFGSvpro. 
In the SA methods, we selected the regularization parameters from GNvpro and varied the batch size and learning rate such that Nesterov had a good validation accuracy (see~\cref{fig:hyperparameterTuningNesterov} for learning rate examples) 
We used the same parameters for ADAM. 
We note that while our experiments utilize the best parameters for GNvpro and Nesterov, we tested the other parameter options on all optimization methods. 
The best parameters for GNvpro and Nesterov corresponded to one of the top performances of the other methods as well; there was not another set of parameters we tested that considerably improved the presented results in~\cref{fig:cifarConvergenceAccuracy}.
}

\end{document}